\documentclass[11pt,draftcls,onecolumn,journal]{IEEEtran}

\usepackage{amsmath, amssymb, amsthm}
\newtheorem{theorem}{Theorem}

\usepackage{graphicx}
\usepackage{caption}
\usepackage{hyperref}
\usepackage{subcaption}
\usepackage{tikz}
\usepackage{wrapfig}
\usepackage[dvipsnames]{xcolor}

\usetikzlibrary{topaths, shapes ,arrows ,positioning}
\tikzset{bend angle=15,
}
\newtoks\mycoords
\usepackage{xparse,fp,xfp}
\usetikzlibrary{decorations.pathreplacing}
\usepackage{microtype}
\usepackage{graphicx}

\usepackage{pgfplots}
\usetikzlibrary{calc}
\usepackage{tikz-3dplot}
\tdplotsetmaincoords{70}{110}
\usetikzlibrary{arrows.meta}

\usepackage{algorithm}
\usepackage{algorithmic}
\usepackage{xcolor}
\usepackage[nodisplayskipstretch]{setspace}
\usepackage[sort&compress,numbers]{natbib}
\usepackage[capitalize,noabbrev]{cleveref}

\usepackage[many]{tcolorbox}    	
\usepackage{setspace}               
\usepackage{multicol}               
\definecolor{main}{HTML}{5989cf}    
\definecolor{sub}{HTML}{cde4ff}     

\tcbset{
    sharp corners,
    colback = white,
    before skip = 0.2cm,    
    after skip = 0.5cm      
}                           
\newtcolorbox{boxC}{
    colback = sub, 
    boxrule = 0pt  
}

\newtcolorbox{boxB}{
    fontupper = \bf\color{main}, 
    boxrule = 1.5pt,
    colframe = black, 
    rounded corners,
    arc = 5pt   
}

\usetikzlibrary{shapes,arrows}

\newcommand{\lasso}{Lasso} 
\newcommand{\params}{\theta} 
 
\newcommand{\Wul}[2]{\mathbf{W}^{\left(#1,#2\right)}} 
\newcommand{\sWul}[2]{w^{\left(#1,#2\right)}} 
\newcommand{\tWul}[2]

\newcommand{\wl}[1]{\mathbf{W}^{\left({#1}\right)}}

\newcommand{\nc}{non-convex}
\newcommand{\nd}[1]{{#1}{-}D}

\newcommand{\hyperbox}{Hyperplane Arrangement Patterns}

\newcommand{\zonotopebox}{Hyperplane Arrangements and Zonotopes}
\newcommand{\samplingbox}{Sampling Arrangement Patterns}
\newcommand{\scalingbox}{Optimal Neural Scaling}
\newcommand{\dualitybox}{Convex Duality of $2$-layer ReLU Networks}

\newcommand{\zonotopeboxref}{``\zonotopebox{}"}
\newcommand{\samplingboxref}{``\samplingbox{}"}

\newcommand{\dualityboxref}{``\dualitybox{}"}
\newcommand{\hyper}{\mathcal{H}}
\newcommand{\revision}[1]{\textcolor{black}{#1}}
\newcommand{\minorrevision}[1]{{#1}}
\newcommand{\finalrevision}[1]{{#1}}

\newcommand{\nonzero}{\revision{non-zero{}} }

\begin{document}

\title{Unveiling Hidden Convexity in Deep Learning: \\A Sparse Signal Processing Perspective\\
}

\author{Emi Zeger and Mert Pilanci\\Department of Electrical Engineering, Stanford University\footnote{© 2026 IEEE. Personal use of this material is permitted. Permission from IEEE must be obtained for all other uses, in any current or future media, including reprinting/republishing this material for advertising or promotional purposes, creating new collective works, for resale or redistribution to servers or lists, or reuse of any copyrighted component of this work in other works.}}


\maketitle

    \section*{Introduction}
    
    Deep neural networks (DNNs), particularly those \minorrevision{using} Rectified Linear Unit (ReLU) activation functions, have achieved remarkable success across diverse machine learning tasks, including image recognition, audio processing\minorrevision{,} and language modeling \cite{lecun2015deep}. Despite this success, the non-convex nature of \minorrevision{DNN} loss functions complicates optimization and \minorrevision{limits} theoretical understanding. These challenges are especially pertinent for signal processing applications where stability, robustness, and interpretability are crucial.
    
    \qquad Various theoretical approaches have been developed for analyzing neural networks \minorrevision{(NNs)}. In this paper, we
highlight \minorrevision{how} recently developed convex equivalence\minorrevision{s} of ReLU \minorrevision{NN}s and their connection\minorrevision{s} to sparse signal processing models can address \minorrevision{the challenges of training and understanding NNs.} Recent research has uncovered several hidden \minorrevision{convexities} in the loss landscapes of certain \minorrevision{NN} architectures, notably two-layer ReLU networks \minorrevision{and other deeper or varied} architectures. 
    By reframing the training process as a convex optimization task, it becomes possible to efficiently find globally optimal solutions. \minorrevision{This approach} offers new perspectives on the network’s generalization and robustness characteristics while facilitating interpretability. Leveraging \lasso{}\minorrevision{-type models} and structure-inducing regularization frameworks, which are fundamental tools in \minorrevision{sparse representation modeling} and compressed sensing \minorrevision{(CS)}, \minorrevision{NN} training can be approached as a convex optimization problem, enabling the interpretation of both globally and locally optimal solutions. 

    \qquad This paper \minorrevision{seeks to} provide an accessible and educational overview that bridges recent advances in the mathematics of deep learning with traditional signal processing, \minorrevision{encouraging}
    broader \minorrevision{signal processing} applications. 
    The paper is organized as follows. We first give a brief background on \minorrevision{NN}s and approaches to \minorrevision{analyzing} them using convex optimization. We then \minorrevision{present} an equivalence theorem between a \minorrevision{two}-layer ReLU \minorrevision{NN} and a convex group \lasso{} problem. We describe how deeper networks and alternative architectures can also be formulated as convex problems, \minorrevision{discuss geometric insights}, and \minorrevision{present} experimental results demonstrat\minorrevision{ing} performance benefits of training \minorrevision{NN}s as convex model\minorrevision{s}. Finally, we discuss remaining challenges and research directions for convex analysis of \minorrevision{NN}s. 


    \minorrevision{
    \textbf{Related tutorial literature}: A number of recent works touch on aspects of regularized neural network training from a signal processing perspective. One such line of work, by \citet{Parhi2023}, studies neural weight balancing, analyzing how parameter magnitudes are distributed under a fixed total $\ell_2$ norm constraint. While this viewpoint is related to the convex duality formulations referenced here, the present article adopts a broader perspective that brings together multiple convex formulations and highlights their implications across a range of neural network models and learning settings. A distinct and complementary direction, explored in \citet{SunSPM}, investigates how regularization and the addition of special neurons can make the loss landscape benign, where local minima are globally optimal. While regularization is also key to improving the loss landscape in the present paper, we focus on approaches to guarantee that the NN loss is not just benign, but convex.
    } 

    \textbf{Notation}: \minorrevision{D}enote $\mathbf{1}$ \minorrevision{as the vector of ones and $[n]=\{1,\cdots,n\}$}. The \minorrevision{B}oolean function $1\{x\}$ returns $1$ if $x$ is true, and $0$ otherwise. Functions and operations \minorrevision{such as} ${1}\{x\}$, $\geq$, \minorrevision{and activations $\sigma(x)$} \minorrevision{apply element-wise} to vector inputs. The $d \times d$ identity matrix is $\mathbf{I}_d$. There are \minorrevision{$m$ hidden neurons and} $n$ training samples $\mathbf{x}\finalrevision{_{(i)}}\finalrevision{\in\mathbb{R}^d}$, which are stacked column-wise into the data matrix $\mathbf{X}\in\mathbb{R}^{d\times n}$ of rank $r$. \minorrevision{NNs $f:\mathbb{R}^d \to \mathbb{R}$ are defined for \finalrevision{column} vectors and extend to data matrix inputs as  $f(\mathbf{X})=(f(\mathbf{x}\finalrevision{_{(1)}}), \cdots, f(\mathbf{x}\finalrevision{_{(n)}}))^T$.}

    \section*{Background: Neural networks}
    NNs are parameterized functions used 
    \minorrevision{for} supervised learning, \minorrevision{composed of functions called} neurons:
    \begin{equation}
    \label{eq:neuron}
     f_{\text{neuron}}(\mathbf{x})=\sigma(\mathbf{x}^T\mathbf{w})
    \end{equation}
    \minorrevision{mapping a vector input to a scalar},
    where $\mathbf{w}\in\mathbb{R}^{d}$ 
    is a \emph{weight}
    parameter
    and $\sigma:\mathbb{R}\to\mathbb{R}$ is \minorrevision{a} nonlinear \emph{activation} function. A common activation function is the Rectified Linear Unit (ReLU): $\sigma(x)=\max\{x,0\}$. An activation function is \emph{active} at ${x}$ if $\sigma(x)\neq 0$. \revision{The activation $\sigma$ is (positively) \emph{homogeneous} (of degree \minorrevision{one}) if $\sigma(a x)=a\sigma(x)$ for $a>0$ \citep{Haeffele2017}.} 
    For example, non-negative inputs activate a ReLU. The nonlinearity of an activation function distinguishes a neuron from a traditional linear model. The neuron \eqref{eq:neuron} is inspired by a biological neuron in the brain, which receives synaptic inputs (\minorrevision{represented by} $\mathbf{x}$) whose intensit\minorrevision{ies are}  modulated by the number of receptors ($\mathbf{w}$)\minorrevision{,} and then fires action potential\minorrevision{s} as output\minorrevision{s}. In the brain, neurons can operate in series, feedback, and parallel pathways \cite{Luo:2021}. Motivated by \minorrevision{these biological features}, a \emph{hidden layer} \minorrevision{consists} of $m$ parallel neurons: $f_{\text{layer}}:\mathbb{R}^{ d}\to\mathbb{R}^{\minorrevision{1\times }m}$,
    \begin{equation}\label{eq:layer}
    f_{\text{layer}}(\mathbf{x})=\sigma(\mathbf{x}^T\mathbf{W}),
    \end{equation}
     where    $\mathbf{W}\in\mathbb{R}^{d\times m}$ is \minorrevision{called a weight matrix}.
    An $L$-layer \minorrevision{NN} \minorrevision{consists of composing} $L-1$ nonlinear hidden layers \eqref{eq:layer}, followed by a \minorrevision{linear output layer}. \minorrevision{The depth is $L$}, and the width of a layer is the number of neurons in that layer. A standard \minorrevision{two}-layer NN $ f:\mathbb{R}^{ d}\to\mathbb{R}$ \minorrevision{takes the form}
    \begin{equation}\label{eq:2layerNN}
        f(\mathbf{x})=\sigma(\mathbf{x}^T\mathbf{W})\boldsymbol{\alpha}
    \end{equation}
    where \eqref{eq:layer} is the first hidden layer of \eqref{eq:2layerNN}, and $\boldsymbol{\alpha}\in\mathbb{R}^{\revision{m}}
    $ \minorrevision{is} the weight \minorrevision{vector}
    of the outer linear layer. Bias parameters can be added implicitly by appending a $1$ to $\mathbf{x}$ and an extra row to $\mathbf{W}$. 
    A\minorrevision{n L-layer NN} is said to be shallow if \minorrevision{$L=2$} and deep if \minorrevision{$L>2$}. The \minorrevision{NN} \eqref{eq:2layerNN} is a \minorrevision{two}-layer, \emph{fully connected}, \emph{feed-forward} network. There are many types of \minorrevision{NN}s consisting of variations on \eqref{eq:2layerNN}, \minorrevision{as well as} deeper networks. 
     NNs fit data to labels/targets based on known pairs $(\mathbf{x}\finalrevision{_{(1)}},y_1),\cdots,(\mathbf{x}\finalrevision{_{(n)}},y_n)$ that are given, where $y_i \minorrevision{ \in \mathbb{R}}$ is the \emph{target} of training sample $\mathbf{x}\finalrevision{_{(i)}}$. A \minorrevision{NN} is \emph{trained} by finding parameters (weights and biases) so that $f(\mathbf{x}\finalrevision{_{(i)}})\approx y_i$,  formulated as a \emph{training problem}
\begin{equation}\label{eq:general_train}
        \min_{\theta} 
       \ell \left(f\left(\mathbf{X}\right), \mathbf{y}\right) 
        + R(\theta)
    \end{equation}
    \minorrevision{where $\mathbf{X} \in \mathbb{R}^{d \times n}$ contains $\mathbf{x}\finalrevision{_{(i)}}$ as columns, $\mathbf{y}\in \mathbb{R}^n$ contains $y_i$}, $\ell$ is a \revision{convex} loss function quantifying the error of \minorrevision{the NN fit}, 
    \minorrevision{and} $\theta$ is the \minorrevision{parameter set, e.g., $\theta=\{\mathbf{W},\boldsymbol{\alpha}\}$ for two-layer NNs \eqref{eq:layer}.} $R$ is a regularization function that penalizes large parameter magnitudes, \minorrevision{encouraging simpler solutions, and will be specified for each NN.} \minorrevision{The training problem \eqref{eq:general_train} is \nc{} due to the product of inner and outer weights. This poses a major challenge for analyzing NNs. Traditional approaches to \revision{train} NNs involve gradient descent algorithms, but these methods can converge to suboptimal local minima. In contrast, if the training loss is convex, all stationary points would be globally optimal. The next section discusses approaches to improve the loss landscape.}
\minorrevision{
\subsection*{Improving the Loss Landscape}
Beyond reducing training complexity, regularization can also improve the geometry of the training objective, or \emph{loss landscape}. In this context, a common goal is to enforce \emph{benign} behavior, meaning that optimization methods are less likely to encounter undesirable stationary points. A representative example, surveyed in \cite{SunSPM} and references therein, shows that augmenting a network with additional ``special'' neurons together with appropriate regularization can eliminate suboptimal local minima (under suitable assumptions). These ideas extend beyond binary classification to multiclass and regression settings, and illustrate how architectural modifications and regularization can jointly shape the loss landscape. We refer the reader to \cite{SunSPM} for precise statements and conditions.
}
%
%
\revision{\subsection*{Computational Complexity}}
\revision{\minorrevision{A well-known result is that the global optimization of two-layer ReLU NNs requires exponential complexity in the input dimension $d$ (see e.g. \cite{froese2023training} and the references therein). A relatively less known result is that, even for fixed $d$,} 
training
is NP-hard, implying no known polynomial-time algorithm exists unless P = NP \cite{froese2023training}. However, this hardness result applies specifically to \minorrevision{unregularized training}. In contrast, as discussed \minorrevision{below}, convex formulations \minorrevision{for networks} incorporating \minorrevision{various} regularization\minorrevision{s} can circumvent NP-hardness, achieving polynomial-time complexity for fixed \minorrevision{$d$} \cite{pilanci2020neural}.} 
\revision{
\minorrevision{While} the \minorrevision{convex} formulations remain computationally demanding in high-dimensional settings,  randomized approximation schemes \cite{Bai2022} related to \minorrevision{1-bit compressed sensing \cite{Boufounos2008} can guarantee} near-global optimality. \minorrevision{In addition to convex approximations, exact convex formulations for NNs enjoy close connections to sparse signal processing. The next section reviews some of these signal processing methods.}
} \\
%
%
%
%
%
    \section*{Background on Sparse Signal Processing, Compressed Sensing, and Geometric Algebra}

Sparse signal processing \minorrevision{analyzes signals that admit a sparse representation} in some basis or dictionary. \minorrevision{Such} sparsity \minorrevision{enables} efficient storage, transmission, and processing of signals. A key application of sparse signal processing is CS  \cite{candes2006robust, donoho2006compressed}, \minorrevision{which studies how} sparse signals can be recovered from far fewer measurements than traditionally required by Nyquist sampling theory. \minorrevision{Various algorithms such as matching pursuit achieve sparsity; we focus on \lasso{}-like models for their connection to \minorrevision{NNs}.}

\subsection*{\lasso}\label{section:lasso} 
The Least Absolute Shrinkage and Selection Operator (\lasso) \cite{tibshirani1996regression} is a convex model \minorrevision{widely used} in sparse signal processing. \lasso{} performs variable selection and regularization simultaneously, making it an effective tool for \minorrevision{sparse representation} in \minorrevision{CS} applications.
The \lasso{} optimization problem is:
\begin{equation}\label{eq:lasso}
\min_{\mathbf{z}}  \frac{1}{2} \| \mathbf{A} \mathbf{z} - \mathbf{y} \|^2_2 + \beta \|\mathbf{z}\|_1
\end{equation}
 where \( \mathbf{A}\) is the measurement \minorrevision{(or dictionary)} matrix whose columns $\mathbf{A}\finalrevision{_{(i)}} \in \mathbb{R}^n$ are called feature vectors or dictionary atoms, \( \mathbf{y} \in \mathbb{R}^n \) is the observation vector, \( \mathbf{z} \in \mathbb{R}^d \) 
 and \( \beta > 0 \) is a regularization parameter controlling the sparsity\minorrevision{, i.e., the number of non-zero elements in a solution for $\mathbf{z}$}. 
 The \( \ell_1 \)-norm \( \|\mathbf{z}\|_1 \) \minorrevision{induces a sparse support of $\mathbf{z}$} by penalizing the absolute sum of the coefficients. 
 In \minorrevision{CS}, \lasso{} serves as a convex relaxation of the NP-hard \( \ell_0 \)-norm minimization problem, providing an efficient computational \minorrevision{framework for inducing sparse representations and variable selection} with theoretical guarantees under certain conditions on the measurement matrix \( \mathbf{A} \). 
%


\subsection*{Group \lasso}
Group \lasso{} \minorrevision{generalizes} \lasso{} to promote \minorrevision{sparse representations} at the group level \cite{yuan2006model}. In many signal processing applications, signals exhibit group sparsity, where \nonzero coefficients occur in clusters. Group \lasso{} accounts for this structure by grouping variables and \minorrevision{using} an \( \ell_{1}/\ell_{2} \)-norm regularization:
\begin{align}
\min_{\mathbf{z}} \frac{1}{2} \|\mathbf{A}\mathbf{z} 
- \mathbf{y}\|_2^2 + \beta \sum_{g=1}^G \|\mathbf{z}\finalrevision{_{(g)}}\|_2\label{eq:grouplasso}
\end{align}
where \( \mathbf{z}\finalrevision{_{(g)}} \) denotes the coefficients in group \( g \), \minorrevision{$G$ is the number of groups}, and \( \|\mathbf{z}\finalrevision{_{(g)}}\|_2 \) is the \( \ell_2 \)-norm of the group coefficients.  This encourages entire groups of coefficients to be zero, promoting structured sparsity that aligns with the underlying signal characteristics. 
When \minorrevision{each group contains a single coefficient $z_i$, the term} $\sum_{g=1}^G \|\mathbf{z}\finalrevision{_{(g)}}\|_2=\sum_{j}|z_j|$ reduces to the $\ell_1$-norm\minorrevision{, recovering the standard \lasso{}. \lasso{} and group \lasso{} models are used in CS, described next.}

\subsection*{Compressed Sensing (CS)}
\minorrevision{Let $\mathbf{z}\in\mathbb{R}^d$ be an unknown signal  
that we can indirectly observe through linear measurements} 
\begin{equation}\label{eq:compsensing}
    y_i=\mathbf{z}^T \mathbf{a}\finalrevision{_{(i)}}
\end{equation}
\minorrevision{
where $\mathbf{a}\finalrevision{_{(i)}}\in\mathbb{R}^d$ are known \emph{measurement vectors}.}
This yields a system of equations $\mathbf{Az}=\mathbf{y}$ where the measurement matrix $\mathbf{A} \minorrevision{\in \mathbb{R}^{M \times d}}$ \minorrevision{has rows} $\mathbf{a}^T_{\finalrevision{(i)}}$ and the observation vector $\mathbf{y} \in \mathbb{R}^M$ contains \minorrevision{corresponding observations} $y_i$. 
We choose the \minorrevision{measurement} vectors $\mathbf{a}^T_{\finalrevision{(i)}}$ and the number of measurements $M$. If \minorrevision{$M=d$ and the measurement vectors $\mathbf{a}_{\finalrevision{(i)}}$ are linearly independent}, then $\mathbf{z}=\mathbf{A}^{-1}\mathbf{y}$ \minorrevision{can be uniquely recovered}. However, in many \minorrevision{applications} (such as MRI image reconstruction), 
\minorrevision{each} measurement incurs costs such as latency or financial \minorrevision{expense}. CS estimates $\mathbf{z}$ \minorrevision{from} fewer \minorrevision{than} $d$ measurements, \minorrevision{effectively} compressing the signal by projecting it onto measurement vectors.
The measurement vectors can be random, \minorrevision{e.g.,} $\mathbf{a}\finalrevision{_{(i)}}\sim \mathcal{N}(0,\mathbf{I}_d)$ \cite{
donoho2006compressed}. 
With fewer than $d$ measurements, multiple candidate solutions satisfy $\mathbf{Az}=\mathbf{y}$;
CS \revision{selects a \minorrevision{parsimonious} solution} by solving \revision{a \lasso{}} \eqref{eq:lasso} or \revision{group \lasso{}} \eqref{eq:grouplasso} depending on the sparsity structure of \minorrevision{$\mathbf{z}$}.
A \minorrevision{variant}, $1$-bit compressive sensing \cite{Boufounos2008}, uses quantized measurements $y_i=Q(\mathbf{z}^T \mathbf{a}\finalrevision{_{(i)}})$, where $Q(x)=\text{sign}(x) \in \{-1,1\}$, compressing each measurement to only one bit of information. 
%
Solving
   \begin{equation*}
       \begin{aligned}
           \min_{\mathbf{z}} &~ \|\mathbf{z}\|_1 \\
            \mbox{s.t. } &\text{Diag}(\mathbf{y}) (\mathbf{A}\mathbf{z} ) \geq 0,~ \|\mathbf{z}\|_2=1
       \end{aligned}
   \end{equation*}
\minorrevision{recovers $\mathbf{z}$},
where the $\ell_2$ constraint resolve\minorrevision{s} ambiguity in the solution. Both CS \minorrevision{and 1-bit CS} models \minorrevision{connect deeply} to convex NNs as we will show in \minorrevision{future sections}.
%
\finalrevision{
Atomic norm theory \cite{atomicnorm} 
views $\mathbf{a}\finalrevision{_{(i)}} \in \mathcal{A}$ as ``atoms," where $\mathcal{A}$ is the set of candidate atoms, 
and chooses $z_i$ by minimizing the Minkowski functional (gauge) of the convex hull of $\mathcal{A}$ as $
\inf\{t>0:\mathbf{y}\in t\,\mathrm{conv}(\mathcal{A})\}$. This defines the atomic norm when $\mathcal{A}$ is symmetric ($-\mathbf{a}_{(i)} \in \mathcal{A}$ if $ \mathbf{a}_{(i)} \in \mathcal{A}$) and reduces to minimizing $\|\mathbf{z}\|_1$ if $\mathcal{A}$ is finite. Convex NN formulations \cite{bengio2005convex,bach2017breaking,zeger2024library,pilanci2020neural} that will be discussed fit into an atomic norm perspective that views neuron outputs as atoms 
and outer-layer weights as  coefficients $z_i$. 
Convex formulations use  $\ell_1$ or group $\ell_1$ penalties, which represent atomic norm regularization that promotes a finite or sparse set of active neurons.
}
\subsection*{Nuclear Norm Regularization}\label{section:nuclearnorm}
The \emph{nuclear norm} of a matrix $\mathbf{W}$ is the $\ell_1$-norm of its singular values: $\|\mathbf{W}\|_*=\sum_{k=1}^{\revision{r}} \sigma_k(\mathbf{W})=\|\sigma(\mathbf{W})\|_1$, \revision{where $r$ is the rank of $\mathbf{W}$} \cite{RechtNuclearNorm}. As a special case, the nuclear norm of a positive semidefinite matrix is its trace. The nuclear norm is often used in optimization problems to search for low-rank matrices \revision{\cite{RechtNuclearNorm}}. We discuss two examples, robust PCA and matrix completion. 

\subsubsection{Robust PCA}
The robust Principal Component Analysis (PCA) problem \finalrevision{\cite{Candes:2011}} for a matrix $\mathbf{X}$ is
\begin{align}\label{eq:robustPCA}
\min_{\mathbf{W},\mathbf{S}:\mathbf{X}=\mathbf{S}+\mathbf{W}} \|\mathbf{W}\|_* + \beta\|\mathbf{S}\|_1. 
\end{align}
The robust PCA problem is a convex heuristic to decompose $\mathbf{X}$ into the sum of a low\revision{-}rank matrix $\mathbf{W}$ and a sparse matrix $\mathbf{S}$ by penalizing the singular values of $\mathbf{W}$ and all elements of $\mathbf{S}$. The low-rank matrix $\mathbf{W}$ can represent the underlying low-dimensional subspace and the sparse matrix $\mathbf{S}$ represents outliers.

\subsubsection{Matrix completion}

The matrix completion problem \finalrevision{\cite{Candes:2010}} is as follows. Given a $n \times n$ matrix where only some of the values are known, fill in the rest of the matrix so that it has the lowest rank possible, consistent with the given elements. This problem can be approximately solved by filling in the unknown values that give the lowest nuclear norm \finalrevision{\cite{Candes:2010}}.

\subsection*{Geometric Algebra}
Geometric \finalrevision{(Clifford)} algebra is a mathematical framework that generalizes complex numbers to higher-dimensional vector spaces, including quaternions and hypercomplex numbers. It provides a \minorrevision{unified} way to represent and perform operations on geometric objects \cite{pilancicomplexity}. Recent work \minorrevision{has applied} geometric algebra to develop new \minorrevision{signal processing techniques} for image, audio, and video processing \cite{Valous2024}, \minorrevision{as well as deep learning} by leveraging geometric representations to enable more compact and expressive models \cite{Valous2024}. \minorrevision{The following sections relate signal processing methods to convex optimization of NNs.}
%



\begin{table}
\begin{center}
\begin{tabular}{ c | c | c | c }
 \shortstack{Convexity w.r.t:} & Approach & Motivation/Technique & \shortstack{Relation to SP} \\ 
 \hline \hline
  \shortstack{outer-layer\\weights} \cite{bengio2005convex} & column generation & \shortstack{
  boosting} & matching pursuit \\
  \hline
 parameters \cite{pilanci2020neural} & \shortstack{convex duality, \lasso{}} & \shortstack{ hyperplane arrangements \\ zonotopes} & \shortstack{sparse recovery, 1-bit CS  \cite{Boufounos2008} } \\
 \hline
 \shortstack{probability\\ measures} \cite{bach:2022, mei2018mean} & \shortstack{mean field\\integration} & \shortstack{neural tangent \\ kernel (NTK)} \cite{JacotNTK2018} & kernel methods \cite{Perez2004} \\
 \hline
 \shortstack{network\\output} \cite{Haeffele2017} & \shortstack{shift complexity \\ to regularization} & \shortstack{
 matrix factorization,\\ nuclear norm \cite{RechtNuclearNorm}} & \shortstack{CS: robust PCA \cite{Candes:2010}\\ matrix completion \cite{Candes:2011} }
 \\
 \hline 
 \shortstack{quantized\\weights} \cite{BIENSTOCK2023} & linear program & \shortstack{intersection \\graph analysis} & \shortstack{graphical models} \cite{wainwright2008graphical} \\
\end{tabular}
\end{center}
    \caption{\minorrevision{Comparison of convex perspectives on NNs. Each approach is motivated by, or involves, techniques related to signal processing (SP). The approaches in the first two rows also involve $\ell_1$ regularizations,  linking them to sparse representations in signal processing.} }
    \label{fig:table}
\end{table}

\section*{\minorrevision{A convex view of \revision{shallow} neural networks}}
    Since convex functions \minorrevision{have been well-studied} \minorrevision{theoretically}, one major approach to study \minorrevision{NN}s \minorrevision{is} to reformulate them via convex optimization. \revision{\minorrevision{Such} convex formulations provide valuable intuition into the underlying structure of optimal network weights and the associated model classes. They enable rigorous analysis of optimality guarantees, principled stopping criteria, numerical stability, and reliability—all critical features for mission-critical applications \minorrevision{common in signal processing}, such as real-time control systems in autonomous vehicles, power grid management, and healthcare monitoring.} 
    \revision{\minorrevision{Moreover}, convex programs \minorrevision{make} training \minorrevision{NN}s agnostic to hyperparameters such as initialization and mini-batching, which \minorrevision{often} exert a significant influence on the performance of local optimization methods. 
    We discuss various approaches to represent \minorrevision{NN} training \minorrevision{problems} as convex problems. This section assume\minorrevision{s} the networks are \minorrevision{two}-layer ReLU networks unless otherwise stated. Deeper networks \minorrevision{follow in} \minorrevision{later} sections.}
\revision{\subsection*{Convexity with Respect to the Outer-Layer Weights}}
\label{sec:cnn-bengio}
\revision{
\citet{bengio2005convex} and \minorrevision{\citet{bach2017breaking}} show that training a two--layer, \minorrevision{infinitely wide NN}
with a convex loss and sparsity-promoting regularization on the outer layer can be cast as a convex
optimization problem. \minorrevision{(In their setup, an $\ell_1$ norm is immediate when the hidden-unit set is countable.)}
Let
\[
\Phi \;=\;\bigl\{\,\sigma(\mathbf{X}^{\mathsf T}\mathbf{w}) :\;
\mathbf{w}\in\mathbb{R}^{d}\bigr\}
\]
denote the (uncountable) set of hidden--layer outputs evaluated on the
training data matrix~$\mathbf{X}\in\mathbb{R}^{d\times \minorrevision{n}}$.
Training reduces to selecting \minorrevision{outer-layer weights}
$\boldsymbol{\alpha}$ by solving \minorrevision{\cite{bengio2005convex, bach2017breaking}}
\begin{equation}
\label{eq:bengio}
\minorrevision{\min_{\boldsymbol{\alpha}}\; \ell\!\Bigl(\int_{\Phi} \phi \, d{\alpha}(\phi), \mathbf{y}\Bigr) +\|{\alpha}\|_{\mathrm{TV}}},
\end{equation}
where $\ell$ is any convex loss,
\minorrevision{the integral integrates over all possible $\phi \in \Phi \subset \mathbb{R}^n$, the final-layer weight $\alpha: \Phi \to \mathbb{R}$ is a finite signed \finalrevision{Radon} measure on $\Phi$ so that $\int_{\Phi}\phi\,d{\alpha}(\phi)\in\mathbb{R}^{n}$ is well-defined, and the regularization is the total variation norm \cite{bach2017breaking}. 
}
\minorrevision{While \cite{bach2017breaking} develops a function-space analysis of \eqref{eq:bengio} by viewing the outer-layer weights as a signed measure over a continuous feature family and regularizing its total variation, the earlier work \cite{bengio2005convex} highlights that the sparsity penalty admits an optimal solution supported on finitely many hidden units (at most $n+1$, by their Theorem~3.1). In this case, the NN reduces to a discrete and finite linear combination of features $\phi$, the continuous distribution $\alpha(\phi)$ in \eqref{eq:bengio} is identified with a vector of coefficients $\boldsymbol{\alpha} \in \mathbb{R}^m$, and the regularization  $\|\boldsymbol{\alpha}\|_{\mathrm{TV}}$ simplifies to the usual $\ell_1$-norm penalty.
}}
\revision{
\citet{bengio2005convex} adopts a
\emph{column generation} strategy to optimize this NN: begin with an empty hidden layer and
iteratively
\emph{(i)}~identify the neuron that most improves the objective and \minorrevision{add it to the hidden layer}  
\emph{(ii)}~re‐optimize~\eqref{eq:bengio} over the current
coefficients. 
Step~\minorrevision{\emph{(i)}} \minorrevision{consists of} solving a weighted linear classification problem,
which is non‐convex in general.  
With hinge loss, the procedure is shown to \minorrevision{reach}
\minorrevision{a} global optimum in at most $n+1$ iterations, corresponding to $n+1$ neurons \cite{bengio2005convex}.  
This incremental scheme links \minorrevision{NN}s to \emph{boosting}---both build predictors by sequentially adding simple components---and, more broadly, to \emph{greedy approximation} methods in signal processing (e.g., matching pursuit and related procedures) \cite{Buhlmann2002}. The extension to continuous distributions in ~\citep{bach2017breaking} also connects naturally to \emph{Frank--Wolfe} style incremental methods.}
\revision{The main computational bottleneck is the non‐convex subproblem used to
insert each neuron, which remains challenging in high dimension\minorrevision{s}.}
\revision{
\minorrevision{From a signal processing perspective,} the features \minorrevision{$\phi \in \Phi$} can be viewed as basis functions for the network \citep{bach2017breaking}.
\minorrevision{The approach in} \citep{bach2017breaking} \minorrevision{is used to analyze}  generalization properties of \minorrevision{infinite-width,} \minorrevision{two}-layer \minorrevision{NNs} with homogeneous activations. 
}
\revision{\subsection*{Convexity with Respect to the Network Output}}
\revision{
In this section we review the convex formulation introduced by \citet{Haeffele2017}.  
Their central idea is to treat the \emph{entire network output} as the decision variable, which converts the learning task into a convex optimization problem.  
Accordingly, the two–layer network in~\eqref{eq:2layerNN} can be rewritten as
\begin{equation}\label{eq:2layerNN_par}
        f(\mathbf{x})=\sum_{i=1}^m \sigma\left(\mathbf{x}^T \mathbf{w}\finalrevision{_{(i)}}\right)\alpha_i 
\end{equation}
\minorrevision{where $\mathbf{w}\finalrevision{_{(i)}} \in \mathbb{R}^d$ is the $i^{\text{th}}$ column of weight matrix $\mathbf{W} \in \mathbb{R}^{d \times m}$ \eqref{eq:layer}.}
%
For a wide class of regularization functions $\minorrevision{R}(\params)$ including norm penalties, the training problem \eqref{eq:general_train} is equivalent to the convex problem 
\begin{equation}\label{eq:haefele_train_vec_form}
        \min_{\mathbf{F} \in \mathbb{R}^n} \ell \left(\minorrevision{\mathbf{F}}, \mathbf{y}\right) + \tilde{R}(\minorrevision{\mathbf{F}})
\end{equation}
where $\tilde{R}(F)$ is the minimum \minorrevision{parameter} regularization such that the corresponding network outputs $\mathbf{F}$ \minorrevision{\citep{Haeffele2017}}:
\begin{equation}\label{eq:aug_reg}
    \tilde{R}(\mathbf{F}) = \min_{\params} R(\params) \text{ s.t. } f(\mathbf{X}) = \minorrevision{\mathbf{F}}
\end{equation}
The \minorrevision{modified} regularization $\tilde{R}$ \minorrevision{\eqref{eq:aug_reg}} is convex \cite{Haeffele2017}, making \eqref{eq:haefele_train_vec_form} convex. In contrast to \cite{Haeffele2017}, the convex problem \eqref{eq:haefele_train_vec_form} is finite-dimensional, as the optimization variable $\mathbf{F}$ is finite-dimensional. This convex reformulation shifts the complexity from optimizing over all weights to optimizing the regularization term.  
\minorrevision{
The approach in \cite{Haeffele2017} is motivated by matrix factorization, as follows.
 Suppose $\mathbf{A} \in \mathbb{R}^{d_1 \times r}, \mathbf{B} \in \mathbb{R}^{r \times d_2}, \mathbf{Y} \in \mathbb{R}^{d_1 \times d_2}$, where $\mathbf{Y}, d_1, d_2$ are known and $\mathbf{A}, \mathbf{B}, r$ are unknown. The matrix factorization problem 
 \begin{equation*}
     \min_{r, \mathbf{A}, \mathbf{B}} \frac{1}{2}\|\mathbf{AB}-\mathbf{Y}\|^2_F+\beta(\|\mathbf{A}\|^2_F + \|\mathbf{B}\|^2_F)
 \end{equation*}
  where the loss function between the matrices is $\ell(\mathbf{Z},\mathbf{Y})=\frac{1}{2}\|\mathbf{Z}-\mathbf{Y}\|^2_F$ and the regularization is $R(\mathbf{A},\mathbf{B})=\beta(\|\mathbf{A}\|^2_F+\|\mathbf{B}\|^2_F)$, can be analyzed similarly to \eqref{eq:haefele_train_vec_form} as
  \begin{equation*}
      \min_{\mathbf{F} \in \mathbb{R}^{d_1 \times d_2}} \frac{1}{2} \|\mathbf{F}- \mathbf{Y}\|^2_F + \tilde{R}(\minorrevision{\mathbf{F}}) 
  \end{equation*}
  where similarly to \eqref{eq:aug_reg}, the modified regularization is
  \begin{equation*}
      \tilde{R}(\mathbf{F})=\min_{r,\mathbf{A},\mathbf{B}} \beta(\|\mathbf{A}\|^2_F + \|\mathbf{B}\|^2_F) \text{ s.t. } \mathbf{AB}=\mathbf{F}
  \end{equation*}
  For matrix factorization, this modified regularization turns out to be the nuclear norm: $\tilde{R}(\mathbf{F})= \|\mathbf{F}\|_*$, a regularization employed in sparse signal processing (see \emph{Nuclear Norm Regularization} above) \cite{Haeffele2017}.
}
However, \minorrevision{for general NNs}, the definition of $\tilde{R}(F)$ \eqref{eq:aug_reg} is not explicit and leaves \eqref{eq:haefele_train_vec_form} intractable.
\minorrevision{Despite this limitation}, convexity \minorrevision{enables useful theoretical insights}. \minorrevision{Consider an augmented training problem where the network is expanded to include one more neuron.} A local \minorrevision{optimum} of the original training problem is globally optimal if adding a neuron with zero-valued weights is locally optimal in the \minorrevision{augmented} training problem \cite{Haeffele2017}. \minorrevision{In particular,} the loss landscape is benign \cite{SunSPM} 
if for some $i\in[m]$, the $i^{\text{th}}$ parameters are zero: $\mathbf{w}\finalrevision{_{(i)}}=\mathbf{0},\alpha_i=0$ in \eqref{eq:2layerNN_par} \cite{Haeffele2017}.  \minorrevision{This perspective on benign landscapes via the addition of neurons is related to the approaches discussed in \cite{SunSPM}: see \emph{Improving the Loss Landscape} above. }}

\revision{\minorrevision{Building off of the benign landscape results,} \citep{Haeffele2017} describes an algorithm to find a globally optimal \minorrevision{NN} by repeating a local descent step that finds a local \minorrevision{optimum} and a\minorrevision{n} expansion step \minorrevision{that adds} another neuron to the \minorrevision{NN}. A local descent path is a sequence of parameter values with a non-increasing objective in the \nc{} training problem. If the network \minorrevision{starts with} at least $n$ neurons (i.e., the number of training samples), a local descent path to a global optimum \minorrevision{exists} \citep{Haeffele2017}. For a sufficiently smooth and strongly convex objective, gradient descent with sufficiently small step size would be an example of local descent \citep{Haeffele2017}. However, for \nc{}
problems, finding a local descent path can be NP-hard \citep{Haeffele2017}. 
In addition, the upper bound on the number of neurons needed to guarantee a local descent path can be large for practical algorithms \citep{Haeffele2017}. A future area of work is to analyze cases where this can be reduced \citep{Haeffele2017}.
}
\revision{\subsection*{Convexity with Respect to Probability Measures}}
\label{sec:meanfield-convex}

\revision{
A complementary route is to lift the \emph{inner} parameters and optimize over a
\emph{probability distribution} on weights (see e.g. \citet{bach:2022, mei2018mean}).  The starting point is to express a finite two-layer network of width $m$ as an
expectation.  If $p_{m}$ denotes the empirical measure that assigns mass
$1/m$ to each parameter pair $(\mathbf{w}\finalrevision{_{(i)}},\alpha_{i})$, then \minorrevision{a $2$-layer \minorrevision{NN} can be represented as} 
\begin{equation}\label{eq:disc_mf}
  f_m(\mathbf{x})
  \;=\;\frac{1}{m}\sum_{i=1}^{m}
        \sigma\!\bigl(\mathbf{x}^{\mathsf T}\mathbf{w}\finalrevision{_{(i)}}\bigr)\alpha_{i}
  \;=\;\mathbb{E}_{(\finalrevision{\boldsymbol{w}},\finalrevision{{\alpha}})\sim p_{m}}
        \!\bigl[\sigma(\mathbf{x}^{\mathsf T} \finalrevision{\boldsymbol{w}})\finalrevision{{\alpha}}\bigr]
\end{equation}
\finalrevision{where $\mathbf{w}_{(i)},\mathbf{w} \in \mathbb{R}^d$ and $\alpha_i, \alpha \in \mathbb{R}$.}
Letting the width $m\to\infty$ while keeping the parameters bounded,
the discrete measures $p_{m}$ converge (along subsequences) to a
probability distribution $p$ on $\mathbb{R}^{d+1}$.  The network output
converges accordingly to the mean-field expression
\begin{equation}
\label{eq:mf-output}
  f_p(\mathbf{x})
  \;=\;\int \sigma(\mathbf{x}^{\mathsf T}\minorrevision{\finalrevision{\mathbf{w}}})\finalrevision{{\alpha}}\,
        \mathrm{d}p(\finalrevision{\mathbf{w}},\finalrevision{{\alpha}}).
\end{equation}
Because $f_p(\mathbf{x})$ depends \emph{linearly} on $p$, any convex loss
$\ell\!\bigl(f(\mathbf{X}),\mathbf{y}\bigr)$ yields a convex training
problem
\[
  \min_{p\in\mathcal{P}}
  \;\ell\!\bigl(f_p(\mathbf{X}),\mathbf{y}\bigr),
\]
over probability measures,
where $\mathcal{P}$ denotes the set of all
non-negative Borel measures on $\mathbb{R}^{d+1}$ whose total mass is
one.
Therefore, the training problem is revealed as a convex optimization in function space.\\
}

\revision{
\emph{Kernel (lazy-training) Regime and the Neural Tangent Kernel}\\
\minorrevision{Let $\phi_m(\mathbf{x}) =
\bigl(
    \sigma(\mathbf{x}^{\mathsf T}\mathbf{w}\finalrevision{_{(1)}}),
    \ldots,
    \sigma(\mathbf{x}^{\mathsf T}\mathbf{w}\finalrevision{_{(m)}})
\bigr)^{\mathsf T}$.}
If \minorrevision{we sample the inner weights i.i.d. from a random distribution, then \emph{freeze} them}
and \minorrevision{optimize} only the outer coefficients, optimizing the \minorrevision{discrete}
mean--field model \eqref{eq:disc_mf} reduces to linear regression:
\minorrevision{$\min_{\boldsymbol{\alpha}} \frac{1}{2}\bigl\|\mathbf{y} - \frac{1}{m}\phi_m(\mathbf{X})^{\mathsf T}\boldsymbol{\alpha}\bigr\|_{2}^{2}$.}
Gradient descent for this problem is known as \emph{kernel gradient descent} with \emph{empirical kernel}
\minorrevision{$\hat{k}_m(\mathbf{x},\mathbf{x}')=\frac{1}{m}\phi_m(\mathbf{x})^{\mathsf T}\phi_m(\mathbf{x}')$.}
As the width \minorrevision{$m$} tends to infinity the empirical kernel
\minorrevision{$\hat{k}_{m}(\mathbf{x},\mathbf{x}')=\frac{1}{m}\sum_{i=1}^{m}
\sigma(\mathbf{x}^{\mathsf T}\mathbf{w}\finalrevision{_{(i)}})\,\sigma(\mathbf{x}'^{\mathsf T}\mathbf{w}\finalrevision{_{(i)}})$}
converges to a deterministic limit \minorrevision{$k_{\mathrm{RF}}(\mathbf{x},\mathbf{x}')$}, \minorrevision{where the subscript `RF' denotes that this limit stems from random features}.
\citet{JacotNTK2018} \minorrevision{show} that if one trains the \emph{full} network but stays in the
\emph{lazy-training} regime, the Jacobian of the loss stays nearly constant and training is equivalent
to kernel gradient descent with a \minorrevision{(generally different but related)} limiting kernel \minorrevision{$k_{\mathrm{NTK}}(\mathbf{x},\mathbf{x}')$}, called the \emph{Neural Tangent Kernel} (NTK).
\minorrevision{The NTK connects NNs to kernel methods, a tool used in signal processing to lift signals
into higher dimensions for linear separation, among other applications \cite{Perez2004}.}
}

    \revision{
    \subsection*{Convexity via \minorrevision{Quantization} and Polyhedral Representations}
    }
    \revision{
    \minorrevision{In another approach,} \minorrevision{\citet{BIENSTOCK2023}} shows that a linear program can approximate \minorrevision{NN} training through weight \minorrevision{quantization} and polyhedral representations. This formulation assumes the training data lies within the box $[-1,1]^d$, \minorrevision{and then} quantizes \minorrevision{the data and weights} to finite numbers of bits. This transforms the training problem into an optimization program with a linear objective, binary variables, and linear constraints, known as a binary program. An \emph{intersection graph} represent\minorrevision{s} the optimization variables \minorrevision{of the binary program} as nodes, and the constraints as edges between them. Relaxing the binary variables to continuous variables converts the binary program into a linear program.
    The treewidth of the intersection graph measures how closely the graph resembles a tree, defined as the minimal width among all possible tree decompositions of the graph. This treewidth characterizes the computational complexity involved in solving the linear program. Specifically, the binary program analyzed in \cite{BIENSTOCK2023} possesses low treewidth, leading to a linear program whose constraints grow linearly with the number of training \minorrevision{samples}. \minorrevision{Therefore, global NN training can be approximated via low treewidth optimization. Treewidth indicates the sparsity of the intersection graph relative to a tree structure, with exact trees having the lowest treewidth. 
    In general, structured sparsity simplifies optimization; in this case, tree-structured sparsity restricts supports to rooted subtrees (connected subgraphs) of a tree. Such signal representations arise in graphical models for image and graph signal processing \cite{wainwright2008graphical}, and decision or regression tree models \cite{Buhlmann2002}. Although these problems are not typically expressed with explicit constraints, they are often formulated as optimization problems that induce implicit intersection graphs capturing parent-child branch dependencies among the signal components. NNs may therefore be interpreted as looser relatives of tree-sparse models with richer parameter interactions
    that still preserve tractability properties associated with low treewidth. }
    In \cite{BIENSTOCK2023}, the polyhedral constraint set captures the \minorrevision{NN} structure.
    Training on different datasets corresponds to solving linear programs constrained by distinct faces of this polyhedron. Importantly, this linear program representation does not require the loss function to be convex. 
    \minorrevision{This approach applies to both unregularized and regularized training, including weight norm penalties as regularization.
    }
    }

\revision{\subsection*{Convexity with Respect to the Network Parameters}}

\revision{The previous sections \minorrevision{describe} approaches \minorrevision{that cast} training problems as implicit convex programs, \minorrevision{either} in terms of the network output or hidden layer \minorrevision{activations}, or by modifying the weights \minorrevision{through} quantization, continuous distributions, or \minorrevision{infinite-width limits}. \minorrevision{In contrast,} a line of recent work \cite{pilanci2020neural,pilancicomplexity} \minorrevision{introduces} explicit convex \minorrevision{formulations that establish} direct correspondences between the variables \minorrevision{of} the convex programs and the network weights. These \minorrevision{formulations involve} a finite number of optimization variables, making them more \minorrevision{representative of} \minorrevision{NN}s in practice. These \minorrevision{convex models} closely parallel \minorrevision{sparse modeling methods in} signal processing such as \lasso{}, group \lasso{}, one-bit CS, and nuclear-norm-based matrix completion.}
This convex approach uncovers certain variable selection properties of weight-decay regularization, \revision{a characteristic of \minorrevision{NN}s also studied in} \citep{bach2017breaking}.
\revision{\minorrevision{Building on these ideas,} \minorrevision{\citet{Bai2022}} develops an adversarial training strategy, \minorrevision{while \citet{prakhya2025convex}} constructs semidefinite relaxations for vector-output, two-layer ReLU networks with $\ell_2$ regularization\minorrevision{.}}
The mathematical proof techniques 
used in the proofs of these results are convex analytic in nature, and include convex geometry, polar duality and analysis of extreme points.
%
We explore the key elements of this convex equivalence, starting with \minorrevision{two}-layer networks.\\
%


\section*{\revision{Explicit Convex Programs for} Two-layer Networks}
\revision{
This section examines two-layer \minorrevision{NNs} with scalar outputs and ReLU activations:
\begin{equation}\label{eq:2lyrReLU}
    f(\mathbf{x}) = \left(\mathbf{x}^T\mathbf{W}\right)_+ \boldsymbol{\alpha},
\end{equation}
and presents an equivalent \lasso{}-type explicit formulation \minorrevision{of the training problem} that establishes a connection between \minorrevision{sparsity-inducing} signal processing and \minorrevision{NN}s. The \lasso{} equivalence gives both practical approaches for training and theoretical insights into \minorrevision{NN} representation power through ideas from signal processing.} 
%
The training problem \eqref{eq:general_train} for the network \eqref{eq:2lyrReLU} using $\minorrevision{\ell}_2$ loss and $\minorrevision{\ell}_2$ regularization is
\begin{align}
%
\min_{\mathbf{W},\boldsymbol{\alpha}} \frac{1}{2}\left \|  f(\mathbf{X}) - \mathbf{y} \right\|_2^2 + \beta \left( 
\|\mathbf{W}\|_F^2 + \|\boldsymbol{\alpha}\|_2^2
\right).
\label{eq:nn}
\end{align}
The term \( \|\mathbf{W}\|_F^2 + \|\boldsymbol{\alpha}\|_2^2 \) represents \emph{weight decay} regularization.
%
%
\minorrevision{We will recast the \nc{} training problem as a convex group \lasso{} problem via hyperplane arrangements, which we define next.}
\subsection*{Hyperplane Arrangement Patterns}

\revision{
A \emph{hyperplane arrangement} is a finite collection of hyperplanes in a given vector space. For simplicity, we first consider hyperplane arrangements in $\mathbb{R}^2$. In  $\mathbb{R}^2$, a \emph{separation pattern} or \emph{hyperplane arrangement pattern} of a dataset is a signed partition of the data by a line through the origin, called a separating line. A separation pattern is represented by a vector $\mathbf{h}\in\{0,1\}^{\minorrevision{n}}$, \minorrevision{where $n$ is the number of training samples}. Specifically, $\mathbf{h}$ is a separation pattern if there exists a separating line such that \minorrevision{each} data point $\mathbf{x}\finalrevision{_{(i)}}$ \minorrevision{is} assigned $h_i=0$ if \minorrevision{it} lies on one side of the line and $h_i=1$ if it lies on the opposite side. Each separating line has an associated orthogonal vector whose orientation determines which side corresponds to $h_i=0$ versus $h_i=1$.
The orthogonal vectors associated with \minorrevision{separating} lines \minorrevision{that give rise to the} same separation pattern \minorrevision{form} a cone, called an \emph{activation chamber}\minorrevision{;} an example is given below. 
\\
}

\begin{figure}
\begin{center}
        \centering
        \setlength{\tabcolsep}{10pt} 
        \renewcommand{\arraystretch}{1.5} 
        \begin{minipage}{0.3\textwidth}
            \begin{tikzpicture}[scale=0.6]
            \draw[] (-2.5,-2) -- (2.5,2) {};
            \draw[]  (-2,1.5) node[fill=black, circle, scale=0.5, label=below:\textcolor{black}{$\mathbf{x}\finalrevision{_{(1)}}$}] () {};
           \draw[]  (1.6,-1.2) node[fill=black, circle, scale=0.5, label=above:\textcolor{red}{$\mathbf{x}\finalrevision{_{(2)}}$}] () {};
           \draw[]  (-1.2,-1.8) node[fill=black, circle, scale=0.5, label=below left:\textcolor{blue}{$\mathbf{x}\finalrevision{_{(3)}}$}] () {};
           \draw[]  (-1.5,-0.5) node[fill=black, circle, scale=0.5, label=left:\textcolor{black}{$\mathbf{x}\finalrevision{_{(4)}}$}] () {};
           \draw[dashed] (-2.5,0) -- (2.5,0) {};
           \draw[dashed] (0,-2.5) -- (0,2.5) {};
           \draw[->] (1.5,1.2) -- (1.9, 0.7) node[below] {${\mathbf{w}}$};
           \filldraw[draw opacity=0, Orchid, fill opacity=0.3] (-2.5,-2.5) -- (-2.5,-2) --(2.5,2) -- (2.5,-2.5) -- cycle;
        \end{tikzpicture}
        \end{minipage}
        \begin{minipage}{0.3\textwidth}
        \quad
     \begin{tikzpicture}[scale=0.6]
           \draw[]  (1.6,-1.2) node[fill=black, circle, scale=0.5, label=above:\textcolor{red}{$\mathbf{x}\finalrevision{_{(2)}}$}] () {};
           \draw[]  (-1.2,-1.8) node[fill=black, circle, scale=0.5, label=left:\textcolor{blue}{$\mathbf{x}\finalrevision{_{(3)}}$}] () {};
           \draw[dashed] (-2.5,0) -- (2.5,0) {};
           \draw[dashed] (0,-2.5) -- (0,2.5) {};
           \filldraw[draw opacity=0, blue, fill opacity=0.2] (-2.5,-2.5) --(2.5,-2.5) -- (2.5, -1.667) -- (-2.5, 1.667) -- cycle;
           \filldraw[draw opacity=0, red, fill opacity=0.2] (2.5,-2.5) -- (2.5,2.5) -- (1.875 , 2.5) -- (-1.875,-2.5) -- cycle;
            \draw[->] (0,0) -- (0.4, -0.5) node[below] {${\mathbf{w}}$};
            \draw[]  (0.7,-2.1) node[] () {\textcolor{Mulberry}{$\mathcal{K}$}};
        \end{tikzpicture}
        \end{minipage}
        \begin{minipage}{0.3\textwidth}
        $\mathbf{h} = \begin{bmatrix}
            0 &
            \textcolor{red}{1} &
            \textcolor{blue}{1}&
            0 
        \end{bmatrix}^T$
        ~\\
        $\mathbf{X} = \begin{bmatrix} \mathbf{x}\finalrevision{_{(1)}} & \textcolor{red}{\mathbf{x}\finalrevision{_{(2)}}} &\textcolor{blue}{ \mathbf{x}\finalrevision{_{(3)}}} & \mathbf{x}\finalrevision{_{(4)}} \end{bmatrix}$
        \end{minipage}%
        \caption{Illustration of a separating hyperplane (left plot) and an activation chamber (right plot).}
        \label{fig:halfspaces}
        \end{center}
        \end{figure}
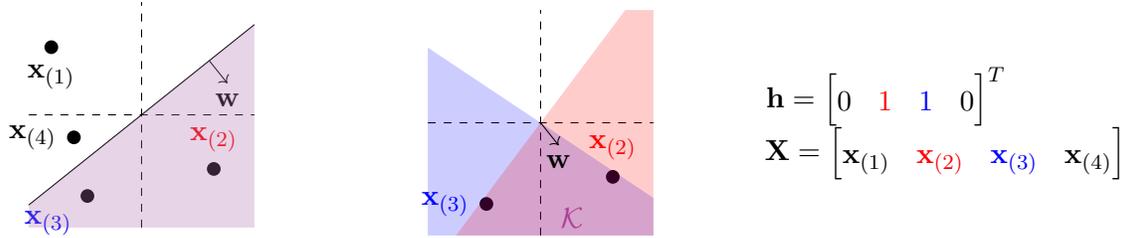

\revision{\textbf{Example}: Consider \nd{2} training data $\mathbf{x}\finalrevision{_{(1)}}, \mathbf{x}\finalrevision{_{(2)}}, \mathbf{x}\finalrevision{_{(3)}}, \mathbf{x}\finalrevision{_{(4)}}$ shown in the left plot of \Cref{fig:halfspaces}. An example of a separation (or hyperplane arrangement) pattern is $\mathbf{h}=(0,1,1,0)^T$. This separation pattern corresponds to $\mathbf{x}\finalrevision{_{(2)}}$ and $\mathbf{x}\finalrevision{_{(3)}}$ being on the same side of \minorrevision{a separating} line \minorrevision{of the form} $\mathbf{w}^T \mathbf{x}=0$. As shown in the figure, $\mathbf{x}\finalrevision{_{(2)}}, \mathbf{x}\finalrevision{_{(3)}}$ \minorrevision{lie} on the side of \minorrevision{this} line in the direction of $\mathbf{w}$ (halfspace shaded in purple), and $\mathbf{x}\finalrevision{_{(1)}},\mathbf{x}\finalrevision{_{(4)}}$ are on the other side. 
In the right plot, the magenta-shaded activation chamber $\mathcal{K}$ \minorrevision{illustrates} the \minorrevision{cone} of all \minorrevision{possible} vectors $\mathbf{w}$ \minorrevision{that are} orthogonal to separating lines 
producing pattern $\mathbf{h}$. 
The red and blue regions are the halfspaces $\mathbf{x}^T_{\finalrevision{(2)}} \mathbf{w} \geq 0$ and $\mathbf{x}^T_{\finalrevision{(3)}} \mathbf{w} \geq 0$, respectively, and the chamber $\mathcal{K}$ is their intersection.}
\revision{The hyperplane arrangement definitions extend to higher dimensions by considering separating hyperplanes instead of separating lines.} The set of \emph{hyperplane arrangement patterns} of a data matrix $\mathbf{X} \in \mathbb{R}^{d\times n}$ is 
\begin{equation}\label{eq:hyperplane_def}
    \hyper=\left\{ {1}\{\mathbf{X}^T\mathbf{w} \geq 0\}: \mathbf{w} \in \minorrevision{\mathbb{R}}^d\right\}.
\end{equation}
Training data on the positive side of the hyperplane $\mathbf{X^T} \mathbf{w}=0$ satisfy $\mathbf{x}^T\mathbf{w}\geq0$ and thus activate the neuron $(\mathbf{x}^T\mathbf{w})_+$, while those on the negative side do not.
%
The number of patterns is 
$|\hyper|\leq2r\left(\frac{e(n-1)}{r}\right)^r$, where $r=\text{rank}(\mathbf{X})\leq\min\{n,d\}$ \revision{and $e=2.718...$ denotes Euler's constant} \cite{pilanci2020neural}. Let $G=|\hyper|$ and enumerate the patterns as $\hyper = \{\mathbf{h}\finalrevision{_{(1)}},\cdots,\mathbf{h}\finalrevision{_{(G)}}\}$. The \emph{activation chamber} 
for a pattern $\mathbf{h}\finalrevision{_{(g)}}\in\mathcal{H}$ is $\mathcal{K}\finalrevision{_{(g)}}=\{\mathbf{w} \in \mathbb{R}^d:{1}\{\mathbf{X}^{\revision{T}}\mathbf{w} \geq \mathbf{0}\}=\mathbf{h}\finalrevision{_{(g)}}\}$, which is the cone of all weights that induce an activation pattern $\mathbf{h}\finalrevision{_{(g)}}$. \finalrevision{Here, $g\in\{1,\cdots,G\}$ is an integer index enumerating the hyperplane arrangement patterns and associated vectors or matrices.}
The \emph{hyperplane encoding matrix} is $\mathbf{D}\finalrevision{_{(g)}}=\text{Diag}(\mathbf{h}\finalrevision{_{(g)}})$ for $g=1,\cdots,G$. For fixed $\mathbf{X}$, a neuron's output $(\mathbf{X}^T\mathbf{w})_+$ as a function of $\mathbf{w}$ is linear over $\mathbf{w}\in\mathcal{K}\finalrevision{_{(g)}}$, as 
$\left( \mathbf{X}^T \mathbf{w} \right)_+=\mathbf{D}\finalrevision{_{(g)}} \mathbf{X}^T \mathbf{w}$. The activation chambers partition $\mathbb{R}^d=\bigcup_{g=1}^G \mathcal{K}\finalrevision{_{(g)}}$, and so the  matrices $\mathbf{D}\finalrevision{_{(1)}},\cdots,\mathbf{D}\finalrevision{_{(G)}}$ completely characterize the piecewise linearity of the neuron output $(\mathbf{X}^T\mathbf{w})_+$. 
\revision{For additional mathematical properties and a thorough introduction to the theory of hyperplane arrangements, we refer the reader to \cite{Stanley2007}.} The training problem is \emph{equivalent} to another optimization problem $P$ if they share the same optimal value, and if an optimal network \minorrevision{to} the training problem can be reconstructed from an optimal solution \minorrevision{to} $P$.
\begin{theorem}[\cite{pilanci2020neural}]\label{theorem:2lyrconvex}
The non-convex training problem \eqref{eq:nn} \minorrevision{with weight decay regularization} for a \minorrevision{two}-layer ReLU network 
is equivalent to the convex group \lasso{} problem
\begin{align}\label{eq:lasso2lyrReLU}
\min_{\mathbf{u}\finalrevision{_{(g)}}, \mathbf{v}\finalrevision{_{(g)}}\in\mathcal{K}\finalrevision{_{(g)}}} \frac{1}{2}\|\sum_{g=1}^G \mathbf{D}\finalrevision{_{(g)}} \mathbf{X }^T\left(\mathbf{u}\finalrevision{_{(g)}} - \mathbf{v}\finalrevision{_{(g)}}\right)- \mathbf{y}\|_2^2 + \beta \sum_{g=1}^G \left(\|\mathbf{u}\finalrevision{_{(g)}}\|_2 + \|\mathbf{v}\finalrevision{_{(g)}}\|_2\right)
\end{align}
where $\mathcal{K}\finalrevision{_{(g)}}=\{\mathbf{z}\finalrevision{_{(g)}}: (2\mathbf{D}\finalrevision{_{(g)}}-\mathbf{I})\mathbf{X}^T\mathbf{z}\finalrevision{_{(g)}}\geq0\}$, provided \revision{the number of neurons satisfies} $m \geq m^*$ where $m^*$ is the number of \nonzero $\mathbf{u}\finalrevision{_{(g)}},\mathbf{v}\finalrevision{_{(g)}}$.
\label{thm:two-layer}
\end{theorem}

\Cref{thm:two-layer} \minorrevision{enables} global optimization of ReLU \minorrevision{NN}s \minorrevision{via} convex optimization and \minorrevision{an} interpretation of the network as a 
signal processing model  \minorrevision{that promotes sparsity, with the dictionary $\mathbf{A}=[\mathbf{D}\finalrevision{_{(1)}}\mathbf{X}^T,...,\mathbf{D}\finalrevision{_{(G)}}\mathbf{X}^T]$.} 
The number $m^*$ is a critical threshold on the number of neurons (number of columns of $\mathbf{W}$) necessary to allow the network to be sufficiently expressive to model the data. 
The \minorrevision{group \lasso{}} variable $\mathbf{z} \in \mathbb{R}^{Gd}$ is partitioned into $G$ consecutive subvectors $\mathbf{z}\finalrevision{_{(g)}}=\mathbf{u}\finalrevision{_{(g)}}- \mathbf{v}\finalrevision{_{(g)}}\in\mathbb{R}^d$, \minorrevision{each corresponding to a single} neuron. \minorrevision{Given an optimal solution to the group \lasso{} problem \eqref{eq:lasso2lyrReLU}}, \minorrevision{the weights for} an optimal \minorrevision{NN} \minorrevision{are} 
\begin{equation}\label{eq:reconstruct}
    \begin{aligned}
        &\mathbf{w}\finalrevision{_{(g)}}=\frac{\mathbf{u}\finalrevision{_{(g)}}}{\sqrt{\|\mathbf{u}\finalrevision{_{(g)}}\|_2}},
        \alpha_g=\sqrt{\|\mathbf{u}\finalrevision{_{(g)}}\|_2} \\
        &\text{ or } \mathbf{w}\finalrevision{_{(g)}}=\frac{\mathbf{v}\finalrevision{_{(g)}}}{\sqrt{\|\mathbf{v}\finalrevision{_{(g)}}\|_2}}, \alpha_g=-\sqrt{\|\mathbf{v}\finalrevision{_{(g)}}\|_2}
    \end{aligned}
\end{equation}
for all non-zero $\mathbf{u}\finalrevision{_{(g)}},\mathbf{v}\finalrevision{_{(g)}}$. \minorrevision{The weight vectors $\mathbf{w}^{(g)}\in\mathbb{R}^d$ are columns of the weight matrix $\mathbf{W}\in\mathbb{R}^{d \times m}$ and $\alpha_g \in \mathbb{R}$ are scalar components of $\boldsymbol{\alpha} \in \mathbb{R}^m$.}  The \lasso{} variables $\mathbf{u}\finalrevision{_{(g)}}$ and $\mathbf{v}\finalrevision{_{(g)}}$ represent weights corresponding to positive versus negative final-layer weights $\alpha_g$. 

    
\textbf{Example} \cite{Ergen2023Landsape}:
Consider the training data matrix
\begin{align}\label{eq:exdata}
    \mathbf{X}=\begin{bmatrix}
    \mathbf{x}\finalrevision{_{(1)}} & \mathbf{x}\finalrevision{_{(2)}} & \mathbf{x}\finalrevision{_{(3)}}
    \end{bmatrix}
    =\begin{bmatrix} 
    2 & 3 & 1\\
    2 & 3 & 0 
    \end{bmatrix}.
\end{align}
Although there are $2^3=8$ distinct binary sequences of length $3$, $\mathbf{X}$ has $G=3$ hyperplane arrangement \revision{patterns} in this case excluding the $0$ matrix, as illustrated below from \cite{Ergen2023Landsape}. \\

%


%
%
\begin{minipage}{.3\textwidth}
\begin{center}
\begin{tikzpicture}[scale=0.7]
\begin{axis}[
  axis lines=middle,
  axis line style={very thick},
  xmin=-6,xmax=6,ymin=-6,ymax=6,
  xtick distance=15,
  ytick distance=15,
  title={},
  grid=major,
  grid style={thin,densely dotted,black!20}]
\addplot [domain=-5:5,samples=2,ultra thick,color=red] {-x*1/3} node[pos=0.2,sloped,above]{linear classifier};
\node[label={90:{(2,2)}},circle,fill,inner sep=2pt] at (axis cs:1,1) {};
\node[label={90:{(3,3)}},circle,fill,inner sep=2pt] at (axis cs:2,2) {};
\node[label={270:{(1,0)}},circle,fill,inner sep=2pt] at (axis cs:1,0) {};
\node[below] at (axis cs:5,0) {$x$};
\node[left] at (axis cs:0,5) {$y$};
\end{axis}
\end{tikzpicture}
\end{center}
\end{minipage}
%
%
\begin{minipage}{.3\textwidth}
\begin{center}
\begin{tikzpicture}[scale=0.7]
\begin{axis}[
  axis lines=middle,
  axis line style={very thick},
  xmin=-6,xmax=6,ymin=-6,ymax=6,
  xtick distance=15,
  ytick distance=15,
  title={},
  grid=major,
  grid style={thin,densely dotted,black!20}]
\addplot [domain=-5:5,samples=2,ultra thick,color=red] {x*1/3} node[pos=0.2,sloped,above]{linear classifier};
\node[label={90:{(2,2)}},circle,fill,inner sep=2pt] at (axis cs:1,1) {};
\node[label={90:{(3,3)}},circle,fill,inner sep=2pt] at (axis cs:2,2) {};
\node[label={270:{(1,0)}},circle,fill,inner sep=2pt,color=red] at (axis cs:1,0) {};
\node[below] at (axis cs:5,0) {$x$};
\node[left] at (axis cs:0,5) {$y$};
\end{axis}
\end{tikzpicture}
\end{center}
\end{minipage}
\begin{minipage}{.3\textwidth}
\begin{center}
\begin{tikzpicture}[scale=0.7]
\begin{axis}[
  axis lines=middle,
  axis line style={very thick},
  xmin=-6,xmax=6,ymin=-6,ymax=6,
  xtick distance=15,
  ytick distance=15,
  title={},
  grid=major,
  grid style={thin,densely dotted,black!20}]
\addplot [domain=-5:5,samples=2,ultra thick,color=red] {x*2/3} node[pos=0.25,sloped,above]{linear classifier};
\node[label={90:{(2,2)}},circle,fill,inner sep=2pt,color=red] at (axis cs:1,1) {};
\node[label={90:{(3,3)}},circle,fill,inner sep=2pt,color=red] at (axis cs:2,2) {};
\node[label={270:{(1,0)}},circle,fill,inner sep=2pt,color=black] at (axis cs:1,0) {};
\node[below] at (axis cs:5,0) {$x$};
\node[left] at (axis cs:0,5) {$y$};
\end{axis}
\end{tikzpicture}
\end{center}
\end{minipage}

\begin{minipage}{.3\textwidth}
\begin{align*}
D_1 =
\left[
\begin{array}{c c c}
1 & 0 & 0\\
0 & 1 & 0\\
0 & 0 & 1
\end{array}
\right]
\end{align*}
\end{minipage}
\begin{minipage}{.3\textwidth}
\begin{align*}
D_2 =
\left[
\begin{array}{c c c}
1 & 0 & 0\\
0 & 1 & 0\\
0 & 0 & \textcolor{red}{0}
\end{array}
\right]
\end{align*}
\end{minipage}
\begin{minipage}{.3\textwidth}
\begin{align*}
D_3 =
\left[
\begin{array}{c c c}
\textcolor{red}{0} & 0 & 0\\
0 & \textcolor{red}{0} & 0\\
0 & 0 & 1
\end{array}
\right]
\end{align*}
\end{minipage}
%
\vspace{1cm}


Consider training a \minorrevision{two}-layer ReLU \minorrevision{NN} on the data matrix in \eqref{eq:exdata}. 
%
For an arbitrary label vector $\mathbf{y} \in \mathbb{R}^3$ and the squared loss, the network in the equivalent convex program \eqref{eq:lasso2lyrReLU} 
%
is written piecewise linearly as
\begin{equation}\label{eq:examplelasso}
    \begin{aligned}
        f(\mathbf{X})
   &=\mathbf{D}\finalrevision{_{(1)}}\mathbf{X}^T(\mathbf{u}\finalrevision{_{(1)}}-\mathbf{v}\finalrevision{_{(1)}})
    +\mathbf{D}\finalrevision{_{(2)}}\mathbf{X}^T(\mathbf{u}\finalrevision{_{(2)}}-\mathbf{v}\finalrevision{_{(2)}})+
    \mathbf{D}\finalrevision{_{(3)}}
    \mathbf{X}^T
    (\mathbf{u}\finalrevision{_{(3)}}-\mathbf{v}\finalrevision{_{(3)}}).
    \end{aligned}
\end{equation}

This \minorrevision{NN} is interpretable \minorrevision{from} a signal processing perspective \cite{yuan2006model}: it looks for a group sparse model to explain the response $\mathbf{y}$ via a mixture of \minorrevision{linear models $\mathbf{u}\finalrevision{_{(g)}}-\mathbf{v}\finalrevision{_{(g)}}$ tuned to each activation chamber.
}
The linear term $\mathbf{u}\finalrevision{_{(2)}}-\mathbf{v}\finalrevision{_{(2)}}$ predicts on $\{ \mathbf{x}\finalrevision{_{(1)}}, \mathbf{x}\finalrevision{_{(2)}}\}$, and $\mathbf{u}\finalrevision{_{(3)}}-\mathbf{v}\finalrevision{_{(3)}}$ on $\{\mathbf{x}\finalrevision{_{(3)}}\}$, etc. Due to the regularization term $\sum_{g=1}^3 \|\mathbf{u}\finalrevision{_{(g)}}\|_2+\|\mathbf{v}\finalrevision{_{(g)}}\|_2$ in \eqref{eq:lasso2lyrReLU}, only a few of these linear terms will be \nonzero at the optimum, showing a bias towards simple solutions among all piecewise linear models. The convex formulation therefore offers insights into the role of regularization in preventing overfitting.  
While the \nd{2} example \eqref{eq:exdata} enumerates hyperplanes by hand, this becomes impractical in high dimensions and is addressed next.
\\

\subsubsection*{\revision{Computational Complexity of Global Optimization and Randomized Sampling for Polynomial-time Approximation}}
\revision{Various approaches have analyzed the complexity resulting from training \minorrevision{NN}s with a convex program.}
The complexity of solving the convex \minorrevision{problem} \revision{\eqref{eq:lasso2lyrReLU}} is proportional to $(\frac{n}{d})^d $, where $n$ is the number of samples and $d$ is the dimension of the input. Although the exponential dependence with respect to $d$ is unavoidable, \minorrevision{the complexity of \eqref{eq:lasso2lyrReLU} is  polynomial in the number of neurons and samples, which is} significantly lower than brute-force search over the linear regions of ReLUs ($O\left(2^{md}\right)$) 
\revision{\cite{pilanci2020neural}. By \minorrevision{leveraging weight decay regularization,} \Cref{thm:two-layer} overcomes the NP-hardness of \minorrevision{unregularized} \minorrevision{NN} training \minorrevision{for fixed $d$} \cite{froese2023training}.  
In \cite{Haeffele2017}, the maximum \minorrevision{width} needed \minorrevision{for} a two-layer \minorrevision{NN} \minorrevision{(trained with a broad class of regularizations)} to reach global optimality corresponds to the number of local descent searches their algorithm must perform, \minorrevision{which} is linear in $n$. However, each local descent step can be NP-hard \cite{Haeffele2017}. }

\Cref{theorem:2lyrconvex} requires a sufficient width of $m\geq m^*$ for the convex equivalence to hold.
\revision{This is consistent with \cite{froese2023training}, \minorrevision{which shows that  training has polynomial time complexity when the number of neurons exceeds a critical width, and can become NP-hard otherwise.} 
The exact width thresholds in \cite{froese2023training} and \Cref{theorem:2lyrconvex} differ because \Cref{theorem:2lyrconvex} considers parameter regularization but \cite{froese2023training} does not.}


\revision{In \cite{BIENSTOCK2023},  polytopes geometrically encode discretized training problems, \minorrevision{both with and without regularization}. While \cite{pilanci2020neural} characterizes a \minorrevision{NN} for fixed training data as an exact piecewise linear model \minorrevision{defined} over a \minorrevision{collection} of polytopes that is exponential in $d$ and polynomial in $\minorrevision{n}$, \cite{BIENSTOCK2023} shows that \minorrevision{a single} polytope, with a number of faces that is exponential in $d$ and linear in $\minorrevision{n}$, \minorrevision{encodes approximations of} training problems for all data sets in $[-1,1]^d$. Specifically, training a network \minorrevision{on} a given data set \minorrevision{amounts} to solving a linear program \minorrevision{over a} face of the polytope, up to an additive quantization error \cite{BIENSTOCK2023}.} 

\begin{boxC}
\textbf{Hyperplane Arrangements and Zonotopes} \\
\begin{wrapfigure}{r}{0.3\textwidth}
\vspace{-2cm}
\begin{center}
\includegraphics[width=0.3\textwidth]{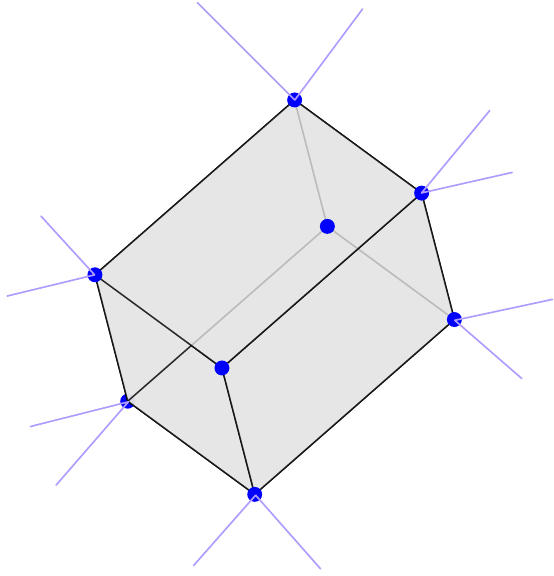}
\caption{Zonotope example. Lines indicate normal cones. \label{fig:zono_with_angles}}
\end{center}
\vspace{-2cm}
\end{wrapfigure}
    Hyperplane arrangements \eqref{eq:hyperplane_def} can be described geometrically. Specifically, the hyperplane arrangement \finalrevision{patterns} of $\mathbf{X}$ correspond to vertices of the \emph{zonotope} (\Cref{fig:zono_with_angles}) of $\mathbf{X}$, defined as 
    \begin{equation}
        \begin{aligned}\label{eq:zonotope}
        \mathcal{Z}&=\text{Conv}\left\{\sum_{i=1}^n h_i \mathbf{x}\finalrevision{_{(i)}}: h_i \in \{0,1\}\right\}
        =\{\mathbf{X} \mathbf{h}: \mathbf{h}\in[0,1]^n\}.
    \end{aligned} 
    \end{equation}

    The correspondence is as follows \cite{Ergen2023Landsape}.
    Since $\mathcal{Z}$ is a polytope, for every $\mathbf{w} \in \mathbb{R}^d$, there is a vertex $\mathbf{z}^*$ of $\mathcal{Z}$ such that 
    \begin{equation}
        \begin{aligned}
            \left(\mathbf{z}^*\right)^T \mathbf{w} = S(\mathbf{w}) = \max_{\mathbf{z} \in \mathcal{Z}} \mathbf{z}^T \mathbf{w}.\label{eq:support}
        \end{aligned}
    \end{equation}
    The line $\mathbf{z}^T \mathbf{w} = (\mathbf{z}^*)^T \mathbf{w}$ is a supporting hyperplane of $\mathcal{Z}$, 
    shown in \Cref{fig:angles} \minorrevision{of \textit{Sampling Arrangement Patterns} below}, 
    and $S(\mathbf{w})$ is the support function of $\mathcal{Z}$. We have shown that vertices maximize the support function of $\mathcal{Z}$.
    Conversely, for every vertex $\mathbf{z}^*$, there is a $\mathbf{w}$ such that that $\mathbf{z}^* = \arg\max_{\mathbf{z}\in\mathcal{Z}}\mathbf{z}^T \mathbf{w}$. Now,
   \eqref{eq:support} is equivalent to
    \begin{equation*}
        \max_{h_i \in [0,1]} \sum_{i=1}^n h_i  \mathbf{x}^T_{\finalrevision{(i)}} \mathbf{w}
    \end{equation*}
    whose solution is a hyperplane arrangement pattern $\mathbf{h}=\mathbf{1}\{\mathbf{X}^T\mathbf{w} \geq 0\}$. Therefore each vertex $\mathbf{z}^*$ corresponds to an activation chamber $\{\mathbf{w}: \text{sign}\left(\mathbf{X}^T \mathbf{w} \right)= \mathbf{h} \} $.
\end{boxC}

\revision{The inset} \samplingboxref{} discusses hyperplane sampling \minorrevision{to reduce computational complexity.} 
\revision{\minorrevision{\citet{Bai2022} presents and analyzes a} procedure to sample a collection of hyperplanes that scales linearly with $n$.
} \minorrevision{As shown in \cite{Bai2022}, \cite{Ergen2023Landsape}, it is possible to control the optimization error resulting from randomly sampling arrangement \finalrevision{patterns} for the \lasso{} problem. In particular, sampling $O(P\log(P))$ random arrangement \finalrevision{patterns} yields a globally optimal solution with high probability, where $P$ is the total number of hyperplane arrangement \finalrevision{patterns} \cite{Ergen2023Landsape}.} In \cite{kim2024convex}, it was proven that the patterns can be randomly subsampled to lower the complexity to $\frac{n}{d} \log n$ with a guaranteed $\sqrt{\log n}$ relative approximation of the objective. This enables fully polynomial-time approximation schemes for the convex \minorrevision{NN} program.

\begin{boxC}
\textbf{Sampling Arrangement Patterns}\\
\begin{wrapfigure}{r}{0.45\textwidth}
\vspace{-0.5cm}
\centering
\begin{tikzpicture}[scale=1.2,transform shape]
    \coordinate (x1) at (2,1);
    \coordinate (x2) at (1,2);
    \coordinate (O) at (0,0);
    \coordinate (x1plusx2) at ($(x1) + (x2)$);
    \coordinate (x1perp) at (-1,2);
    \coordinate (x1plusx2perp) at (3,-1);
    \draw[thick,->] (O) -- (x1);
    \draw[thick,->] (x1) -- (x1plusx2);
    \draw[thick] (x2) -- (x1plusx2);
    \draw[thick,->] (O) -- (x2);
    \node at (O) [left] {0};
    \node at (x1) [right] {$x_1$};
    \node at (x2) [left] {$x_2$};
    \node at (x1plusx2) [right] {$x_1+x_2$};
    \coordinate (normal1) at (1,-1.5);  
    \node at ($(x1) + 0.5*(normal1)$) [right] {$h$};
    \coordinate (normal1) at (1,-1.5);  
    \coordinate (normal2) at (-1,3); 
    \coordinate (normal1perp) at (1.5,1);  
    \coordinate (normal2perp) at (3,1); 
    \draw[thick,dashed,blue] ($(x1)!-2.5cm!($(x1)+(normal1perp)$)$) -- ($(x1)!2.5cm!($(x1)+(normal1perp)$)$);
    \draw[->,red,thick] (x1) -- ($(x1) + 0.5*(normal1)$);
    \fill[blue, opacity=0.2] (x1) -- ($(x1) - 1*(x1perp)$) -- ($(x1) + 1*(x1plusx2perp)$) -- cycle;
    \end{tikzpicture}
    \caption{Zonotope normal cones}
    \label{fig:angles}
\end{wrapfigure}
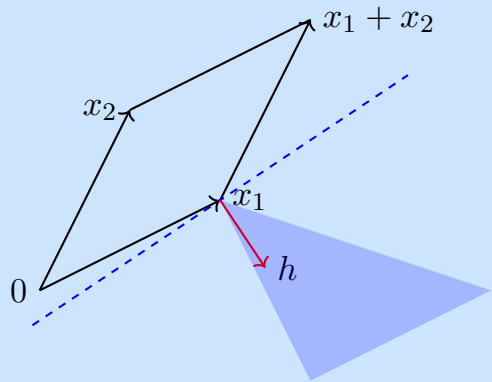
\vspace{-0.3cm}
~\\
    Using a subset of hyperplane arrangement \finalrevision{patterns} \eqref{eq:hyperplane_def} in the \lasso{} problem \eqref{eq:lasso2lyrReLU} can reduce the exponential complexity of enumerating all patterns. This yields a subsampled \lasso{} problem, and is proven to correspond to stationary points
    %
  of the \nc{} training problem
 \cite{Ergen2023Landsape}. 
    %
    \minorrevision{A practical way to sample hyperplane arrangement \finalrevision{patterns} is to generate}
    random $\mathbf{w}\sim \mathcal{N}(0, \mathbf{I}_d)$ and evaluate $\mathbf{h}\finalrevision{_{(g)}}= 1\{\mathbf{X}^T \mathbf{w} \ge 0\}$ \minorrevision{\cite{pilanci2020neural},\cite{Bai2022}}. \minorrevision{Then $\mathbf{D}\finalrevision{_{(g)}}=\text{Diag}(\mathbf{h}\finalrevision{_{(g)}})$ are used to solve an estimated group \lasso{} problem \eqref{eq:lasso2lyrReLU}. An estimated optimal network is reconstructed from the estimated \lasso{} solution via \eqref{eq:reconstruct}. }
    
%
    \minorrevision{
    }%
    The \minorrevision{hyperplane sampling} implicitly samples vertices of the data's zonotope, whose normal cone solid angles (\Cref{fig:angles}) are proportional to the probability of sampling each vertex \cite{Ergen2023Landsape}. \revision{Zonotopes are described in \eqref{eq:zonotope} in the inset \zonotopeboxref. Zonotopes \minorrevision{have various signal processing} applications \cite{Stinson2016}.}

   The hyperplane sampling approach is directly connected to $1$-bit compressive sensing  \cite{Boufounos2008}, which randomly samples a signal $\mathbf{x}\in\mathbb{R}^d$ as $ \text{sign}(\mathbf{x}^T \mathbf{w}\finalrevision{_{(i)}}) $ where $\mathbf{w}\finalrevision{_{(i)}} = \mathbf{a}\finalrevision{_{(i)}}, \mathbf{x}=\mathbf{z}$ in \eqref{eq:compsensing}. 
   Multiple signals stacked into a data matrix $\mathbf{X}$ can be randomly sampled at once as $\text{sign}\left(\mathbf{X}^T \mathbf{w}\right)$ where $\mathbf{w}\sim \mathcal{N}(0,\mathbf{I}_d)$, which is precisely sampling activation chambers, or equivalently, the vertices of a zonotope.
    Sampling patterns is also related to Locality-Sensitive Hashing (LSH), a method for efficiently finding nearest neighbors (in Euclidean distance) of an entry $\mathbf{x}\in \mathbb{R}^d$ in a database \cite{Slaney2008LSH}. 
    LSH places $\mathbf{x} $ in a database bin according to the sign of $\mathbf{x}^T \mathbf{w}$ where $\mathbf{w} \sim \mathcal{N}(0,\mathbf{I}_d)$, similar to 1-bit CS.  

\end{boxC}

\minorrevision{The 
group \lasso{} formulation in \Cref{theorem:2lyrconvex} 
enables model pruning algorithms to generate minimal models \cite{mishkin2023optimal}. A minimal model is a minimum-width, globally optimal \minorrevision{NN}. Starting an optimal convex solution \eqref{eq:lasso2lyrReLU}, \cite{mishkin2023optimal} successively removes 'redundant' neurons corresponding to group \lasso{} variables $\mathbf{u}\finalrevision{_{(g)}}$ whose contributions $\mathbf{D}\finalrevision{_{(g)}} \mathbf{X }^T \mathbf{u}\finalrevision{_{(g)}}$ to the prediction are linear combinations of that from other group variables. The algorithm also extends to constructing approximate minimal models that use fewer neurons than required for global optimality \cite{mishkin2023optimal}.}
%
%
%
%
%
%
The proof of \Cref{thm:two-layer} involves showing that the bidual of the non-convex training problem is equivalent to the \lasso{} problem and reconstructing a network from the \lasso{} problem that achieves the same objective in the training problem as the \lasso{} problem, thus closing the duality gap and proving equivalence. 
\minorrevision{We note that there are alternative ways to obtain the convex formulation. One approach involves showing that neurons \finalrevision{with} the same arrangement \finalrevision{pattern} can be merged and then optimizing over neurons constrained to the activation chambers. Here, we focus on the convex duality approach, which parallels the duality theory of traditional \lasso{} methods used in signal processing.} 
\revision{The next section describes a neural weight property related to the proof. \\}

\textbf{Gaining Insight into Network Weights: \scalingbox{}}.
     \minorrevision{
     \citet{Yang2022ABW} and \citet{Parhi2023} describe a ``Neural \finalrevision{B}alance Theorem'' for training with weight decay regularization. The theorem} states that the magnitude of the input and output weights of any neuron with homogeneous activation must be equal, satisfying a ``balancing constraint." Specifically, the minimum-weight representation of a given function by a two-layer \minorrevision{NN} satisfies $|\alpha_j|=\|\mathbf{w}\finalrevision{_{(j)}}\|_{\minorrevision{2}}$ for all $j$ \finalrevision{where $\mathbf{w}_{(j)}$ if the $j^{\text{th}}$ column of $\mathbf{W}$} \eqref{eq:2layerNN}. The network must have a homogeneous activation such as ReLU. This weight scaling strategy is used as an important first step in the convexification approach of \cite{pilanci2020neural}. 
The ReLU \minorrevision{NN} \eqref{eq:2layerNN_par} $f(\mathbf{x}) = \sum_{j=1}^m (\mathbf{x}^T\mathbf{w}\finalrevision{_{(j)}})_+ \alpha_j$ is invariant to multiplying $\mathbf{w}\finalrevision{_{(j)}}$ and dividing $\alpha_j$ by any positive scalar $\gamma_j$. On the other hand, for fixed $\mathbf{w}\finalrevision{_{(j)}},\alpha_j$, their training regularization $\|\gamma_j \mathbf{w}\finalrevision{_{(j)}}\|^2_2 + |\frac{1}{\gamma_j}\alpha_j|^2$ is minimized over $\gamma_j$ when $\|\gamma_j \mathbf{w}\finalrevision{_{(j)}}\|_2=|\frac{1}{\gamma_j}\alpha_j|$, i.e., $\gamma^*_j = \frac{|\alpha_j|}{\|\mathbf{w}\finalrevision{_{(j)}}\|_2}$. Therefore an optimal \minorrevision{NN} that minimizes the regularized training problem will have equal magnitude inner and outer weights 
$   |\alpha_j|=\|\mathbf{w}\finalrevision{_{(j)}}\|_2 $
 for all $j$. Intuitively, this means that an optimal network balances weights evenly between layers. Similar scaling properties hold for deeper networks and other activations. \revision{Next we discuss the derivation of the explicit convex program \eqref{eq:lasso2lyrReLU} for ReLU NNs using duality theory. \\}
 %

\label{box:derivation}
\textbf{Convex duality of two-layer ReLU networks}.
\revision{The previous discussion} 
shows that the \minorrevision{value of the weight decay} regularization \minorrevision{with optimal parameters} in the training problem \eqref{eq:nn} is \minorrevision{equivalent to} $\beta \sum_{j=1}^m \|\mathbf{w}\finalrevision{_{(j)}}\|_2|\alpha_j| $ and the network can be written as $f(\mathbf{x}) = \sum_{j=1}^m (\mathbf{x}^T\frac{\mathbf{w}\finalrevision{_{(j)}}}{\|\mathbf{w}\finalrevision{_{(j)}}\|_2})_+\alpha_j$. Subsume $\|\mathbf{w}\finalrevision{_{(j)}}\|_2 \alpha_j \to \alpha_j$ and rewrite the training problem \eqref{eq:nn} as
\begin{equation}\label{eq:rescaled}
    \begin{aligned}
        \min_{\|\mathbf{w}\finalrevision{_{(j)}}\|_2=1, \boldsymbol{\alpha}} & \,\, \frac{1}{2}\left\|\mathbf{u} - \mathbf{y}\right\|_2^2 + \beta \|\boldsymbol{\alpha}\|_1 \\
   & \mbox{ s.t. } \mathbf{u}=\sum_{j=1}^m  (\mathbf{\mathbf{X}}^T\mathbf{w}\finalrevision{_{(j)}})_+\alpha_j.
    \end{aligned}
\end{equation}
The Lagrangian of \eqref{eq:rescaled} has linear and $\minorrevision{\ell}_1$ norm terms of $\alpha$, so minimizing it over $\alpha$ gives the dual of  \eqref{eq:rescaled}:
\begin{equation}\label{eq:dual}
    \begin{aligned}
        \max_{\mathbf{v}} & -\frac{1}{2}\|\mathbf{v}-\mathbf{y}\|^2_2 \\
        & \mbox{s.t. } \max_{\|\mathbf{w}\|_2\leq1} 
        | \mathbf{v}^T (\mathbf{\mathbf{X}}^T\mathbf{w})_+|\leq \beta.
    \end{aligned}
\end{equation}
While \eqref{eq:dual} has a semi-infinite constraint over all $\|\mathbf{w}\|_2\leq1$, there are in fact a finite number of unique possible vectors $\left(\mathbf{\mathbf{X}}^T\mathbf{w}\right)_+$, corresponding to activation chambers $\mathcal{K}\finalrevision{_{(g)}}$, as shown in ``\hyperbox{}." Maximizing over each $\mathcal{K}\finalrevision{_{(g)}}$ makes the constraint in  \eqref{eq:rescaled} finite and linear:
\begin{equation}\label{eq:dualfinite}
    \begin{aligned}
        \max_{\mathbf{v}} &\, -\frac{1}{2}\|\mathbf{v}-\mathbf{y}\|^2_2 \\
        & \mbox{s.t. } \max_{\|\mathbf{w}\|_2\leq1, \mathbf{w}\in\mathcal{K}\finalrevision{_{(g)}}} 
        |\mathbf{v}^T \mathbf{D}\finalrevision{_{(g)}} \mathbf{\mathbf{X}}^T\mathbf{w}| 
        \leq \beta \mbox{ for all } g=1,\cdots,G
    \end{aligned}
\end{equation}
Taking the dual of the \eqref{eq:dualfinite}, assigning $\mathbf{u}\finalrevision{_{(g)}},\mathbf{v}\finalrevision{_{(g)}}$ to correspond to positive and negative signs inside the absolute value, and simplifying gives the \lasso{} problem. The \lasso{} problem therefore \minorrevision{presents} a lower bound on the training problem. However, the reconstruction \eqref{eq:reconstruct} gives a network that achieves the same objective, and thus the \lasso{} problem and training problem are equivalent. 
\textbf{Hyperplane enumeration gives the key equivalence of \eqref{eq:dual} and \eqref{eq:dualfinite}, and the key \lasso{} duality is the equivalence of \eqref{eq:dualfinite} and \eqref{eq:lasso2lyrReLU}}. Most importantly, the active constraints in \eqref{eq:dualfinite} identify exactly which groups are selected at an optimum, and hence which neurons appear in the recovered network. This mirrors the Lasso dual, where constraints that are tight pinpoint the active primal coefficients, equivalently the support of the sparse solution.
Similar approaches \minorrevision{can extend to} \minorrevision{deeper} architectures, discussed next.

\section*{Deeper ReLU Networks \minorrevision{and Convexity}}
While shallow networks \minorrevision{suffice for simpler applications}, deeper networks can capture more complex relationships between data. 
\minorrevision{Deep networks can also admit convex programs depending on the architecture.}

\subsection*{Importance of Parallel Architecture}
We consider two architectures for a deep \minorrevision{NN}. First,
a $L$-layer \emph{standard} network extends the two-layer network \eqref{eq:2layerNN} by composing more layers as $f:\mathbb{R}^d\to\mathbb{R}$,

\begin{equation}\label{eq:std}
        f(\mathbf{x})=\left( f^{(L-1)}\circ f^{(L-2)}\circ\cdots\circ f^{(1)} (\mathbf{x})\right)\finalrevision{^T}\boldsymbol{\alpha}
\end{equation}

where 
$f^{(l)}(\mathbf{x})=\sigma\left(\mathbf{x}^T\wl{l}\right)\finalrevision{^T}$ 
is the $l^{\text{th}}$ layer \eqref{eq:layer} and the trainable parameters are weight matrices $\mathbf{W}^{(l)}$ 
for $l \in [L-1]$ and final layer parameter $\boldsymbol{\alpha}$.
%
%
An alternative architecture is a $L$-layer \emph{parallel network} which \minorrevision{is a sum of $K$ standard networks}
as
$ f: \mathbb{R}^{d} \to \mathbb{R}$, 
    \begin{equation}\label{eq:nn_nonrecursive}
        f(\mathbf{x}) = \sum_{k=1}^{K} \left(f^{(k,L-1)}\circ f^{(k,L-2)} \circ \cdots \circ f^{(k,1)}(\mathbf{x}) \right)\finalrevision{^T}\boldsymbol{\alpha}^{\minorrevision{(k)}} 
    \end{equation}
where 
$f^{(k,l)}(\mathbf{x})=\sigma\left(\mathbf{x}^T\Wul{k}{l}\right)\finalrevision{^T}$
is the $l^{\text{th}}$ layer \eqref{eq:layer} of the $\minorrevision{k}^{\text{th}}$ unit, which consists of $m_l$ neurons.  
%
The trainable parameters are the outermost weights $\boldsymbol{\alpha}^{\minorrevision{(k)}} \in \mathbb{R}^{m_{L\minorrevision{-1}}}$
and the inner weights $ \Wul{\minorrevision{k}}{l} \in \mathbb{R}^{m_{l{-}1}\times m_l}$ 
for each layer $l \in [L-1]$ and unit $\minorrevision{k} \in [\minorrevision{K}]$, where $m_0=d$ \minorrevision{and $m_L=1$} \cite{ergen2023path}. 
The standard \eqref{eq:std} and parallel \eqref{eq:nn_nonrecursive} architectures both appear in literature \cite{wang2023parallel, ergen2023path, bach2017breaking, Haeffele2017}, and \minorrevision{coincide} for two-layer \minorrevision{NNs}. 
%
\dualityboxref{} \minorrevision{shows that \cite{pilanci2020neural} constructs} the  convex \minorrevision{NN} formulation as the bidual of, \minorrevision{and hence a lower bound on,} the \nc{} training problem. This lower bound \minorrevision{is tight} for two-layer networks: strong duality \minorrevision{holds} with no duality gap, so the convex formulation is equivalent to the training problem \cite{wang2023parallel}. \revision{The duality gap is the difference between the optimal value of the original problem and its dual (which is shown to be is equivalent to its bidual) \cite{pilanci2020neural}.} Applying the same duality approach to deeper parallel \minorrevision{NN}s \minorrevision{also produces an}  equivalent convex problem with no duality gap \cite{wang2023parallel}. However, this approach for standard networks, even with a linear activation $\sigma(x)=x$, \minorrevision{produces} a \nonzero duality gap \cite{wang2023parallel}, necessitating parallel architectures to extend this convex analysis for more layers. 

    \revision{
   The linear programs representing discretized \minorrevision{NN} training in \cite{BIENSTOCK2023} hold for multiple layers, although solving the linear programs remains intractable. 
In \citep{Haeffele2017}, the global optimality results for a two-layer \minorrevision{NN} analogously extend to deep parallel networks: a \minorrevision{locally optimal parallel NN} is globally optimal if at least one of the parallel units is zero (all weights in the unit are zero matrices). The training problem is equivalent to a convex program analogous to \eqref{eq:haefele_train_vec_form}, but as in the two-layer case, the convex program is intractable with an implicitly formulated regularization $\tilde{R}(\mathbf{F})$ and cannot be typically solved in polynomial time \citep{Haeffele2017}. 
%
}
\revision{\subsection*{Neural Feature Repopulation}
The mean-field convexification approach \minorrevision{\cite{bach:2022, mei2018mean}} extend\minorrevision{s} to multiple layers in \citep{fang2019convex} through a \emph{neural feature repopulation} strategy which considers each layer as a feature. \minorrevision{The \minorrevision{NNs} use a class of regularizers including $\ell_{1,2}$ regularization, where inner and outer weights incur $\ell_2$ and $\ell_1$ norm penalties respectively.} For a network with one neuron in each layer, let $x^{(l)}$ be the vector-valued output of the neuron in layer $l$. \citep{fang2019convex}. Let $w^{(2)}$ be a function that takes in $\mathbf{w}^{(1)}$ and the second layer output $x^{(2)}$ and outputs the second layer weight $\mathbf{w}^{(2)}$. Let $p^{(1)}$ be the distribution of $\mathbf{w}^{(1)}$.
Every possible output of the second layer $x^{(2)}$ is formed by integrating $x^{(1)}=\mathbf{X}^T \mathbf{w}^{(1)}$ passed through an activation and scaled by the second layer weight \citep{fang2019convex}:
\begin{equation}
    x^{(2)}=\int_{\mathbf{w}^{(1)}} w^{(2)}\left(\mathbf{w}^{(1)},x^{(2)}\right) \sigma\left(x^{(1)}\right) dp^{(1)}(\mathbf{w}^{(1)}).
\end{equation}
For deeper layers, the vector weight $\mathbf{w}^{(l)}$ between layers $l$ and $l+1$ is viewed as the output of a weight function $w^{(l)}$ that takes as input $(x^{(l)}, x^{(l+1)})$ \citep{fang2019convex}. Also let $p^{(l)}$ be the probability distribution of the output of the $l^{\text{th}}$ layer.
Every intermediate layer between the first and last hidden layer satisfies
\begin{equation}
    x^{(l+1)}=\int_{x^{(l)}} w^{(l)}\left(x^{(l)},x^{(l+1)}\right) \sigma\left(x^{(l)}\right) dp^{(l)}(x^{(l)}).
\end{equation}
Finally, the last layer satisfies 
\begin{equation}
    x^{(L+1)}=\int_{x^{(L)}} \alpha\left(x^{(L)}\right) \sigma\left(x^{(L)}\right) dp^{(L)}(x^{(L)})
\end{equation}
where $\alpha$ is the outer weight as a function of the output of the last hidden layer.
The composition of layers is converted to constraints in a convex optimization problem with an objective that is linear in the feature corresponding to the output layer \citep{fang2019convex}. The complexity is confined to the regularization. 
%
While this approach \minorrevision{is effective} for a finite
number of layers
the convex problem is infinite-dimensional (in the number of optimization variables). 
On the other hand, the convex equivalences formulated in \cite{pilanci2020neural} extend\minorrevision{s} to \minorrevision{three}-layer ReLU parallel networks, giving explicit convex training programs as detailed next. 
}

\subsection*{\revision{Explicit \minorrevision{Three}-layer Convex Networks with Path Regularization}}\label{section:3layreReLU}
The convex formulation \eqref{eq:lasso2lyrReLU} extends to deeper networks \cite{ergen2023path}. 
The training problem for a \minorrevision{three}-layer parallel ReLU network $f(\mathbf{x})=\sum_{\minorrevision{k}=1}^{\minorrevision{K}} \left(\left(\mathbf{\minorrevision{x}}^T\Wul{\minorrevision{k}}{1}\right)_+\Wul{\minorrevision{k}}{2}\right)_+\minorrevision{\boldsymbol{\alpha}^{(k)}}$  with \emph{path regularization} \minorrevision{is}
\begin{align}
\min_{\minorrevision{\theta}} \frac{1}{2} \minorrevision{\|f(\mathbf{X})-\mathbf{y}\|^2_2} + \beta \sum_{\minorrevision{k}=1}^{{\minorrevision{K}}} \sqrt{\sum_{\minorrevision{i},j} \left \|\mathbf{w}^{(\minorrevision{k},1)}_{\minorrevision{i}} \right\|^2_2 \left(\sWul{\minorrevision{k}}{2}_{\minorrevision{i},j}\right)^2 \left(\alpha^{\minorrevision{(k)}}_j\right)^2} 
\label{eq:train3lyrReLU}
\end{align}
where \minorrevision{the parameter set is $\theta=\{\mathbf{W}^{(k,1)}, \mathbf{W}^{(k,2)},\boldsymbol{\alpha}^{(k)}: k \in [K]\}$,} $\mathbf{w}^{(k,1)}_i$ is the $i^{\text{th}}$ column of $\Wul{k}{1} \in \mathbb{R}^{d \times m_1}$, $\sWul{k}{2}_{i,j}$ is the $(i,j)^\text{th}$ element of $\Wul{k}{2} \in \mathbb{R}^{m_1 \times m_2}$ \minorrevision{and $\alpha_j^{(k)}$ is the $j^{\text{th}}$ element of $\boldsymbol{\alpha}^{(k)} \in \mathbb{R}^{m_2}$}. The regularization penalizes all of the paths from the input through the parallel networks to the output. The \minorrevision{three}-layer ReLU training problem \eqref{eq:train3lyrReLU} is equivalent to the convex group \lasso{} problem

\begin{align}
\min_{\mathbf{u}_{\finalrevision{(i,j,k)}}, \mathbf{v}_{\finalrevision{(i,j,k)}}\in\mathcal{{K}}_{\finalrevision{(i,j)}}} \frac{1}{2}\Big\|\sum_{i=1}^{G_1}\sum_{j,k=1}^{G_2}\mathbf{D}_{\finalrevision{(i)}} {\mathbf{D}}'_{\finalrevision{(j,k)}} \mathbf{X }^T(\mathbf{u}_{\finalrevision{(i,j,k)}} - \mathbf{v}_{\finalrevision{(i,j,k)}})- \mathbf{y}\Big\|_2^2 + \beta \sum_{i,j,k}\left(\|\mathbf{u}_{\finalrevision{(i,j,k)}}\|_2 + \|\mathbf{v}_{\finalrevision{(i,j,k)}}\|_2\right)
\end{align}
where $\mathbf{D}_{\finalrevision{(i)}}, \mathbf{D}'_{\finalrevision{(j,k)}}$ are diagonal matrices encoding hyperplane arrangement patterns in the first and second layer, and $\minorrevision{\mathcal{K}}_{\finalrevision{(i,j)}}$ are multilayer activation chambers \cite{ergen2023path}. Multilayer hyperplanes divide the two-layer activation chambers $\mathcal{K}_{\finalrevision{(i)}}$ into subchambers $\mathcal{K}_{\finalrevision{(i,j)}}$, partitioning the data into finer regions where the \minorrevision{NN} acts as \minorrevision{a} local linear model. The convex \minorrevision{perspective} shows that \minorrevision{NNs gain representation power with depth by learning activation patterns that capture more complex structures in the data} and tailoring linear models more locally within each activation chamber.  
The convex formulation provides a geometric interpretation of the network operations, revealing the underlying sparsity structure. A network reconstruction formula is given in \cite{ergen2023path}. \minorrevision{For the deep learning community, the convex recasting of training eliminates the need for reducing local minima, determining initialization, and other hyperparameter tuning and heuristics used in training.}
\revision{A limitation of this method is the restriction on network architecture. In deeper models, the number of features increases significantly, making it more difficult to use the convex formulations.}
%
%

\revision{\section*{\minorrevision{Interpretability of neural networks via convexity}}}  

\revision{The previous sections described geometric connections of \minorrevision{NN} training to polyhedron faces \cite{BIENSTOCK2023}, as well as hyperplanes and zonotopes \cite{pilanci2020neural}. Here, we describe a \finalrevision{polytope based} interpretation.
}

\revision{\subsection*{\nd{1} data: features encode distances between data points}
Thus far, the \minorrevision{input} dimension to a \minorrevision{NN} has been arbitrary. However, the special case of \nd{1} data offers concrete insights on \minorrevision{NN} structure. Consider a \minorrevision{two}-layer network with bias parameters $f(\mathbf{x})=\sigma\left(\mathbf{x}^T\mathbf{W}+\mathbf{b}\right)\boldsymbol{\alpha}$ where $\mathbf{b}$ is a $m$-dimensional row vector representing trainable but unregularized bias parameters. \minorrevision{The training problem} for this \minorrevision{two-layer} network with \minorrevision{\nd{1} inputs,} ReLU activation\minorrevision{, and weight decay regularization 
$\beta\left(\|\mathbf{W}\|_F^2 +\|\boldsymbol{\alpha}\|_2^2\right)
$
} is shown in \cite{zeger2024library} to be equivalent to the \lasso{} problem
\begin{equation}\label{eq:\lasso2}
\min_{\mathbf{z}}  \frac{1}{2} \| \mathbf{A}^+ \mathbf{z}^+ + \mathbf{A}^- \mathbf{z}^- - \mathbf{y} \|^2_2 + \beta \|\mathbf{z}\|_1.
\end{equation}
The submatrices of the dictionary \minorrevision{in \eqref{eq:\lasso2}} are $\mathbf{A}^+, \mathbf{A}^- \in \mathbb{R}^{n\times n}$ with  $A^+_{i,j}=\sigma(x\finalrevision{_{(i)}}-x\finalrevision{_{(j)}}), A^-_{i,j}=\sigma(x\finalrevision{_{(j)}}-x\finalrevision{_{(i)}})$. For symmetric activations such as $\sigma(x)=|x|$, we can replace $\mathbf{A}$ by $\mathbf{A}^+$ and $\mathbf{z}$ by $\mathbf{z}^+$. In contrast to the constrained group \lasso{} \eqref{eq:lasso2lyrReLU} which requires enumerating hyperplane arrangement patterns, the convex problem \eqref{eq:\lasso2} is a standard, unconstrained \lasso{} problem \eqref{eq:lasso} with a straightforward dictionary to construct.} \minorrevision{This \lasso{} formulation extends to higher-dimensional data by identifying dictionary matrices $\mathbf{A}^+$ and $\mathbf{A}^-$ 
in 
\eqref{eq:\lasso2} 
with volumes of oriented simplices formed by the data points \cite{pilancicomplexity}, discussed next.}

\subsection*{High dimensional data: geometric algebra features encode volumes spanned by data}

 A key tool in generalizing \lasso{} results to higher dimensions is \finalrevision{geometric} \finalrevision{a}lgebra, \minorrevision{and in particular, its wedge product operation.} \finalrevision{Geometric} \finalrevision{a}lgebra is a mathematical framework that has been recently explored in signal processing \cite{Valous2024}. 
The \emph{wedge product} of vectors $\mathbf{a}$ and $\mathbf{b}$ is denoted as $\mathbf{a}\wedge\mathbf{b}$, which measures the signed area of the parallelogram spanned by $\mathbf{a}$ and $\mathbf{b}$. 
\minorrevision{The wedge product of two \nd{2} vectors $\mathbf{a}=(a_1,a_2)^T$ and $\mathbf{b}=(b_1,b_2)^T$ is the determinant of the matrix they form:
$\mathbf{a} \wedge \mathbf{b} = \text{det}\begin{pmatrix}
    a_1 & b_1 \\
    a_2 & b_2
\end{pmatrix}$ 
= $a_1b_2 - b_1 a_2 = 2\mathbf{Vol}(\triangle(0,\mathbf{a},\mathbf{b}))$, where 
$\triangle(\mathbf{c},\mathbf{a},\mathbf{b})$ is the triangle with vertices $\mathbf{c},\mathbf{a},\mathbf{b}$ and $\mathbf{Vol}$ denotes the 2-volume (area). 
The \minorrevision{two}-layer, univariate-input ReLU dictionary elements \eqref{eq:\lasso2} $A^+_{i,j}=(x_{\finalrevision{(i)}}-x_{(j)})_+$ represent the positive length of
the interval $[x_{\finalrevision{(i)}}, x_{\finalrevision{(j)}}]$. On the other hand \cite{pilancicomplexity}, we can also write
$(x_{\finalrevision{(i)}}-x_{\finalrevision{(j)}})_+ =
\left(\begin{bmatrix}
    x_{\finalrevision{(i)}} \\
    1
\end{bmatrix}
\wedge
\begin{bmatrix}
    x_{\finalrevision{(j)}} \\
    1
\end{bmatrix}\right)_+$, which equals twice the positive area of the triangle with vertices $(x_{\finalrevision{(i)}},1)^T$, $(x_{\finalrevision{(j)}},1)^T$ and the origin. This triangle has unit height and base $x_{\finalrevision{(i)}}-x_{\finalrevision{(j)}}$. Lifting the training points $x_{\finalrevision{(i)}}, x_{\finalrevision{(j)}} \in \mathbb{R}$ to $(x_{\finalrevision{(i)}},1)^T, (x_{\finalrevision{(j)}},1)^T \in \mathbb{R}^2$ corresponds to appending a vector of ones to the data matrix to represent a bias parameter in the network.
}

\begin{wrapfigure}[9]{l}{0.3\textwidth}
\vspace{-0.6cm}
    \begin{tikzpicture}[scale=0.6]
        \def\veci{(4,0,0)}
        \def\vecjone{(1,3,0)}
        \def\vecjtwo{(1,1,0)}
        \draw[fill=blue, opacity=0.3] (0,0)--(4,0)--(5,3)--(1,3)--(0,0);
        \draw[fill=red, opacity=0.3] (5,3)--(1,3)--(2,4)--(6,4)--(5,3);
        \draw[fill=yellow, opacity=0.3] (4,0)--(5,3)--(6,4)--(5,1)--(4,0);
        \draw[dashed, red] (2,4)--(1,1);
        \draw[dashed, red] (0,0)--(1,1);
        \draw[dashed, red] (1,1)--(5,1);
        \draw[->, ultra thick] (0,0,0) -- \veci node[anchor=north east] {\(\mathbf{x}_{\finalrevision{(i)}}\)};
        \draw[->, ultra thick] (0,0,0) -- \vecjone node[anchor=north east] {\(\mathbf{x}_{\finalrevision{(j_1)}}\)};
        \draw[->, ultra thick] (0,0,0) -- \vecjtwo node[anchor=north west] {\(\mathbf{x}_{\finalrevision{(j_2)}}\)};
        \node at (4,4.5,0) {\(\mathbf{x}_{\finalrevision{(i)}} \wedge \mathbf{x}_{\finalrevision{(j_1)}} \wedge \mathbf{x}_{\finalrevision{(j_2)}}\)};
        \node[anchor=north east] at (0,0,0) {$0$};
        \draw[fill=green, opacity=0.3] (0,0)--\veci--\vecjone--cycle; 
        \draw[fill=green, opacity=0.3] (0,0)--\veci--\vecjtwo--cycle; 
        \draw[fill=green, opacity=0.3] (0,0)--\vecjone--\vecjtwo--cycle; 
        \draw[fill=green, opacity=0.3] \veci--\vecjone--\vecjtwo--cycle; 
        \fill (0,0,0) circle (2pt); 
        \fill \veci circle (2pt);
        \fill \vecjone circle (2pt);
        \fill \vecjtwo circle (2pt);
    \end{tikzpicture}
    \caption{Wedge product in geometric algebra \cite{pilancicomplexity}.}
    \label{fig:wedge}
  \end{wrapfigure}
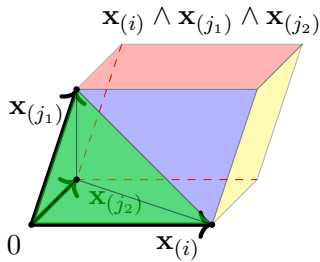

\minorrevision{Now c}onsider a \minorrevision{two}-layer \minorrevision{ReLU} network $f(\mathbf{x})=\sigma(\mathbf{x}^T\mathbf{W})\boldsymbol{\alpha}$ \minorrevision{(without a bias parameter $\mathbf{b}$)} trained on $\mathbf{x}\finalrevision{_{(i)}}\in \mathbb{R}^{2}, y_i\in \mathbb{R}$ \minorrevision{and t}he \nc{} training problem \eqref{eq:general_train} with $\minorrevision{\ell}_2$-norm loss and $\minorrevision{\ell}_p$-norm weight decay regularization \cite{pilancicomplexity} \minorrevision{$\|\mathbf{W}\|^2_p + \|\boldsymbol{\alpha}\|^2_p$}. \minorrevision{Consider a dictionary for the \lasso{} problem \eqref{eq:lasso} with elements} 
 $A_{ij} = \frac{2}{\|\mathbf{x}_{\finalrevision{(j)}}\|_p}\mathbf{Vol}(\triangle(0,\mathbf{x}_{\finalrevision{(i)}},\mathbf{x}_{\finalrevision{(j)}}))_+$
 \cite{pilancicomplexity}. 
 If we add a bias term to the neurons, then \minorrevision{let}
$A_{ij} = \frac{2}{\|\mathbf{x}_{\finalrevision{(j_1)}}-\mathbf{x}_{\finalrevision{(j_2)}}\|_p}\mathbf{Vol}(\triangle(\mathbf{x}_{\finalrevision{(i)}},\mathbf{x}_{\finalrevision{(j_1)}},\mathbf{x}_{\finalrevision{(j_2)}}))_+$ where $j=(j_1,j_2)$ \minorrevision{represents a multi-index}. 
Here, $\mathbf{Vol}(\cdot)_+$ distinguishes triangles based on the orientation, i.e., whether their vertices form clockwise or counter-clockwise loops. The \lasso{} problem \minorrevision{\eqref{eq:lasso} is equivalent to the \nc{} training problem} when the regularization is $\minorrevision{\ell}_1$-norm \minorrevision{($p=1$),} and \minorrevision{produces} an $\epsilon$-optimal network when the regularization is $\minorrevision{\ell}_2$-norm \minorrevision{($p=2$)} \cite{pilancicomplexity}.
\minorrevision{These results generalize to higher dimensions, with wedge products corresponding to higher-dimensional volumes of parallelotopes spanned by vectors.}
\Cref{fig:wedge} from \cite{pilancicomplexity} illustrates a wedge product in $\mathbf{R}^3$ \minorrevision{as the volume of a parallelepiped spanned by $3$ vectors}.
The higher-dimensional dictionary elements are $A_{ij} = \frac{\left( \mathbf{x}_{\finalrevision{(i)}} \wedge  \mathbf{x}_{\finalrevision{(j_1)}} \wedge \dots \wedge  \mathbf{x}_{\finalrevision{(j_{d-1})}} \right)_+}{\|  \mathbf{x}\finalrevision{_{(j_1)}} \wedge \dots \wedge  \mathbf{x}\finalrevision{_{(j_{d-1})}}\|_2},$ 
leading to a convex program that captures the geometric relationships in the data \cite{pilancicomplexity}. 
Notably, $A_{ij}$ is a ratio of volumes of \minorrevision{parallelotopes}, where the positive part of the signed volume encod\minorrevision{es} the order of the indices.
The geometric algebra approach provides a unified framework for interpreting deep \minorrevision{NN}s as a dictionary of wedge product features, revealing the underlying geometric structures inherited from the training data. \minorrevision{The volumes of parallelotopes spanned by the training data determine the features. In \nd{1}, the \minorrevision{two}-layer NN training problem is equivalent to a classical linear spline \lasso{} model where the features are hinge functions. The wedge product dictionary shows that in higher dimensions, these hinge functions generalize to signed distances to affine hulls spanned by training data points. The wedge product formulation also shows that optimal weights are orthogonal to training data \cite{pilancicomplexity}.} 
%
%
%
This result establishes a theoretical foundation for the success of deep \minorrevision{NNs} by linking them to equivalent Lasso formulations whose dictionaries capture geometric and symmetry properties, providing novel insights into the impact of depth on \minorrevision{NN} expressivity and generalization.



\section*{Extensions to Other Neural Network Architectures}

\revision{
The convex approaches described in this paper extend to various architectures beyond \minorrevision{two}-layer ReLU networks.
The convex training with respect to outer-layer weights in \cite{bengio2005convex} generalizes to multiple layers, and the parameter distribution approach in \cite{bach:2022} extends to homogeneous activations. While \cite{bach:2022} qualitatively analyzes training convergence, guarantees on convergence \minorrevision{rate} and the network width are areas of future work. The implicit convex model with respect to \minorrevision{NN} output in \cite{Haeffele2017} also extends to multiple layers and a wide family of homogeneous activations. A limitation in \cite{Haeffele2017} is that a potentially large number of neurons in \minorrevision{two}-layer \minorrevision{NN}s (or parallel units in deeper networks) are required to guarantee global optimality of training, and a future direction is reducing this size. The linear programs in \cite{BIENSTOCK2023} apply to a variety of architectures including deep networks, convolutional networks, residual networks, vector outputs, and activations.
}
The convex representation approach in \cite{pilanci2020neural} extends to modern architectures revealing new forms of convex regularizers. The following contains a brief overview (see \cite{Ergen2023Landsape} and the references therein). \revision{The convex \lasso{} approach extends to \finalrevision{NNs with} polynomial or threshold activations \finalrevision{which are not homogeneous.
For threshold activations, including amplitude parameters ensures the neural balance effect \cite{Yang2022ABW,Parhi2023}. For polynomial activations, regularizing final layer weights $\boldsymbol{\alpha}$ with $\ell_1$ regularization is essential for the associated duality results.} \finalrevision{Convexification also extends to} networks with quantized weights, generative adversarial networks and diffusion models used to produce synthetic images and other data, and}
 Convolutional \minorrevision{NN}s (CNNs), \minorrevision{which} are popular for image and audio \minorrevision{processing} problems. 
Transformers and attention mechanisms are used in Large Language Models for understanding human language, and are convexified by \minorrevision{leveraging} nuclear norm regularization, \minorrevision{a tool used in sparse signal processing.}
Convex reformulation techniques also reveal that batch normalization, an algorithmic trick essential for the success of local search heuristics, corresponds to applying a whitening matrix to the data, \minorrevision{a preprocessing technique used in signal estimation \cite{Felhauer1993}.} 
    %
Next, we discuss numerical experiments.

\begin{figure}[t!]
\begin{subfigure}{0.49\textwidth}
\vspace{-1cm}
    \centering
    \includegraphics[width=1.23\textwidth]{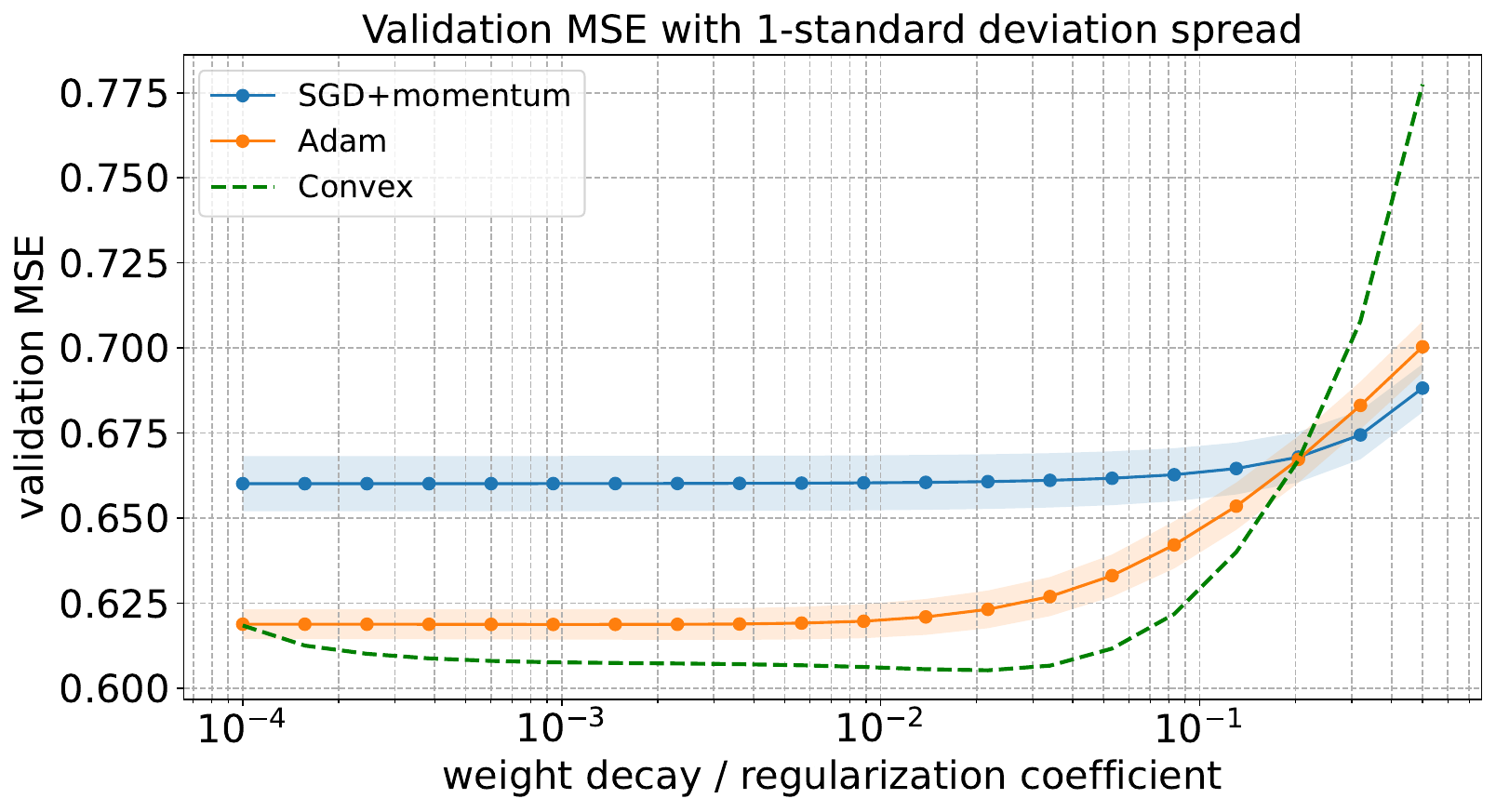} 
    \vspace{-0.55cm}
    \caption{Time series forecasting of NY Stock Exchange.}
    \label{fig:nyse}
\end{subfigure}
\begin{subfigure}{0.5\textwidth}
\hspace{1.5cm}
    \centering
        \includegraphics[width=0.65\textwidth]
        {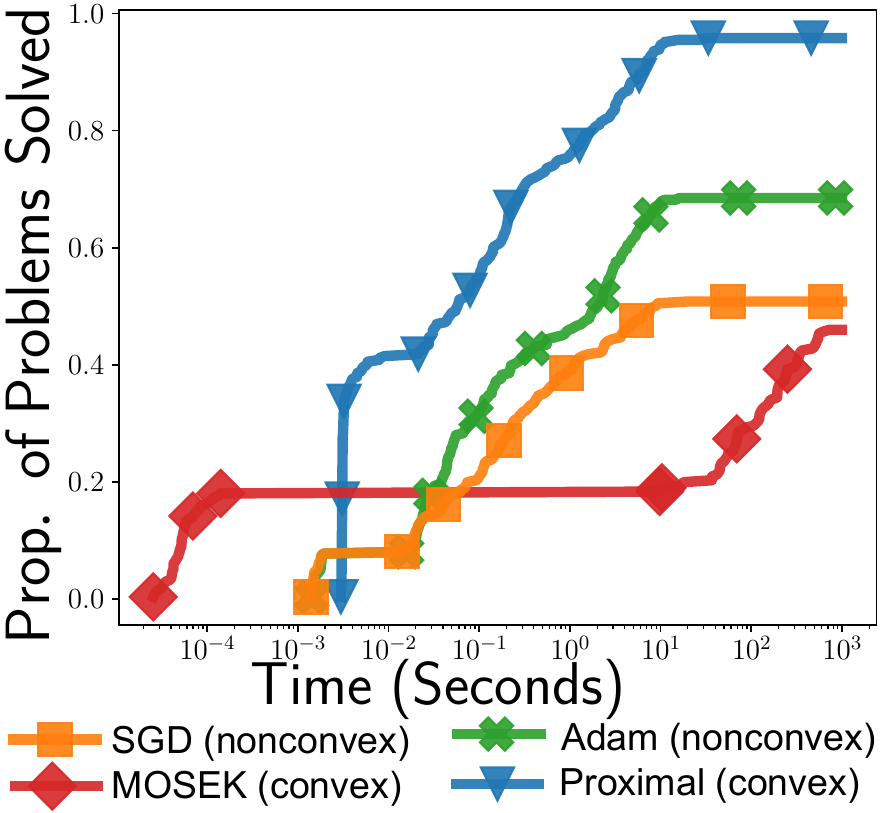}
        \vspace{0cm}
        \caption{Performance profile reproduced from \cite{mishkin2022fast}. }
        \label{fig:convexnn}
\end{subfigure}
\caption{ Experimental results comparing $2$-layer \minorrevision{NN} training with \nc{} and convex formulations.}
\end{figure}

    \section*{Simulations}
\begin{figure}[t]
     \centering
     \begin{subfigure}[t]{0.45\textwidth}
     \includegraphics[width=\textwidth]{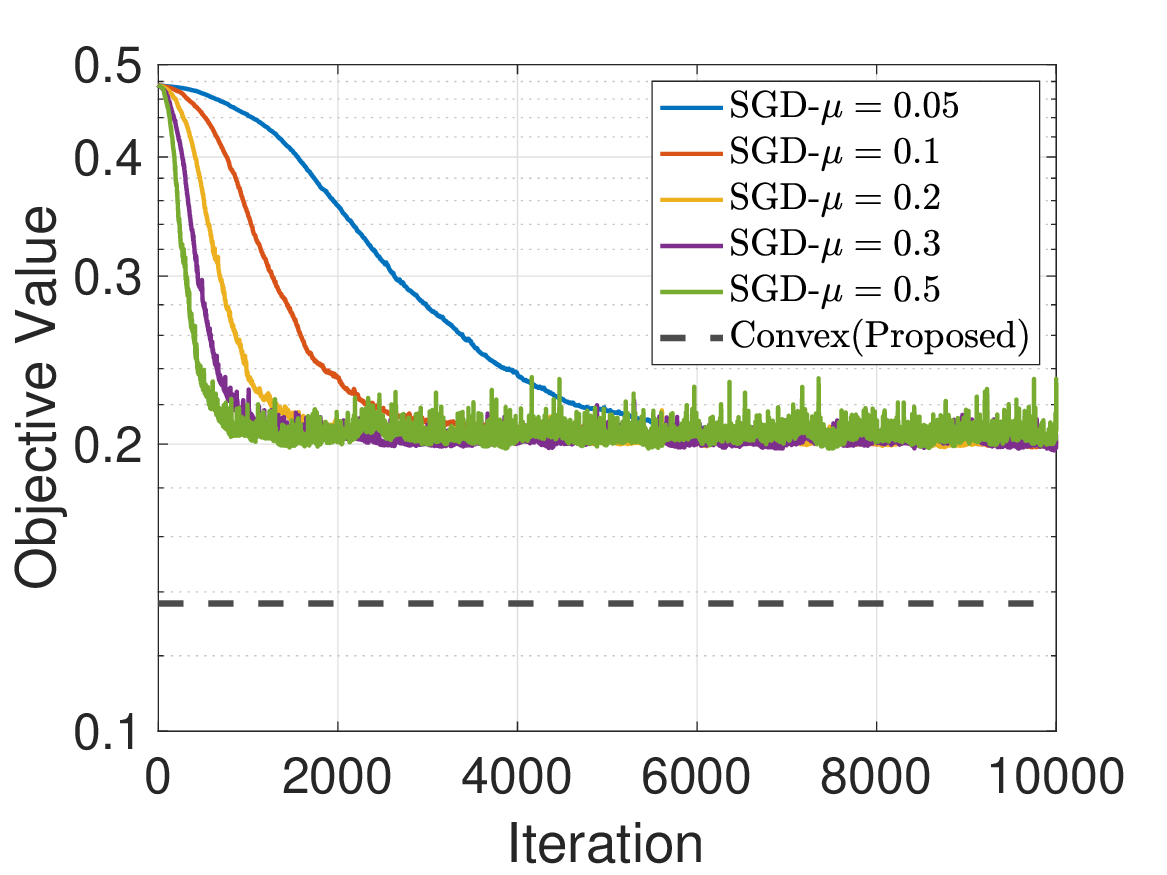} 
     \caption{Training objective value}
     \label{fig:ecg_obj}
     \end{subfigure}
     \hfill
     \begin{subfigure}[t]{0.45\textwidth}
     \includegraphics[width=\textwidth]{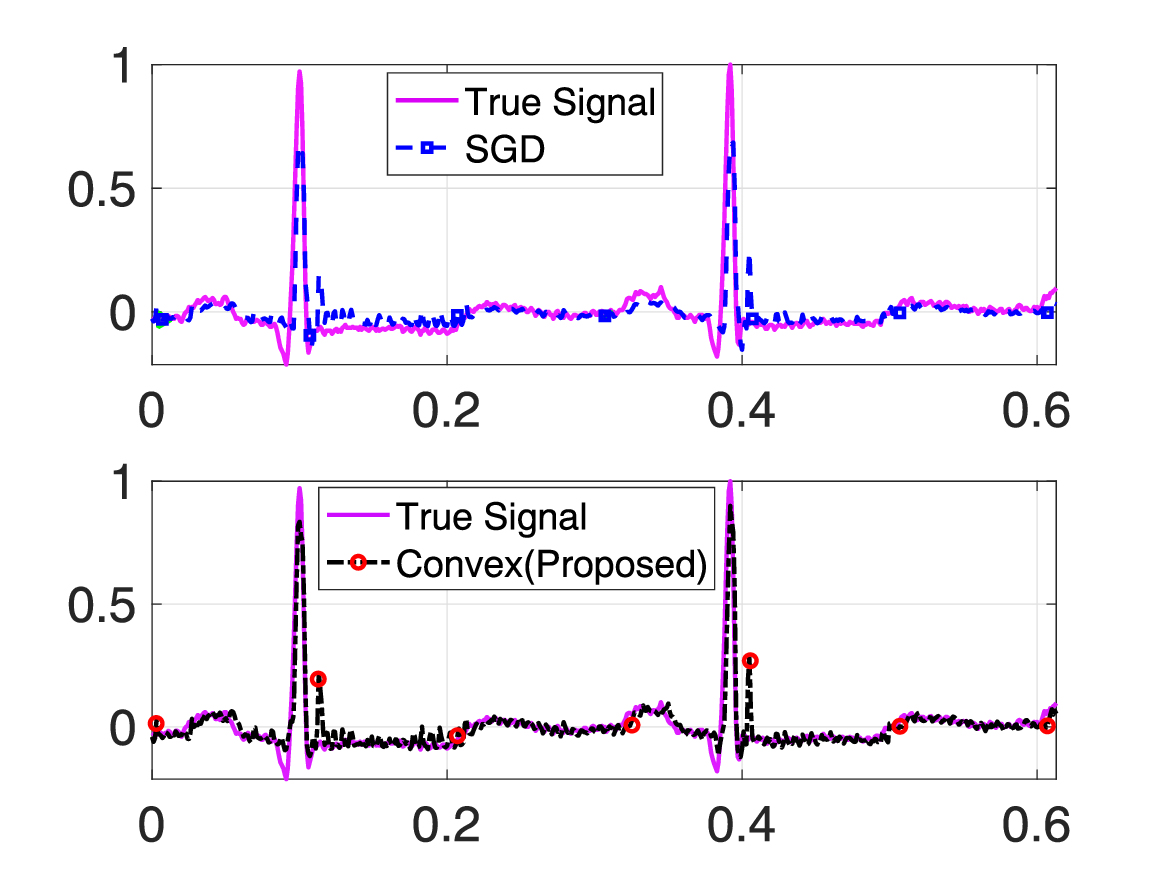} 
     \caption{Test predictions}
     \label{fig:ecg_pred}
     \end{subfigure}
     \begin{subfigure}[t]{0.48\textwidth}
        \includegraphics[width=0.9\textwidth]{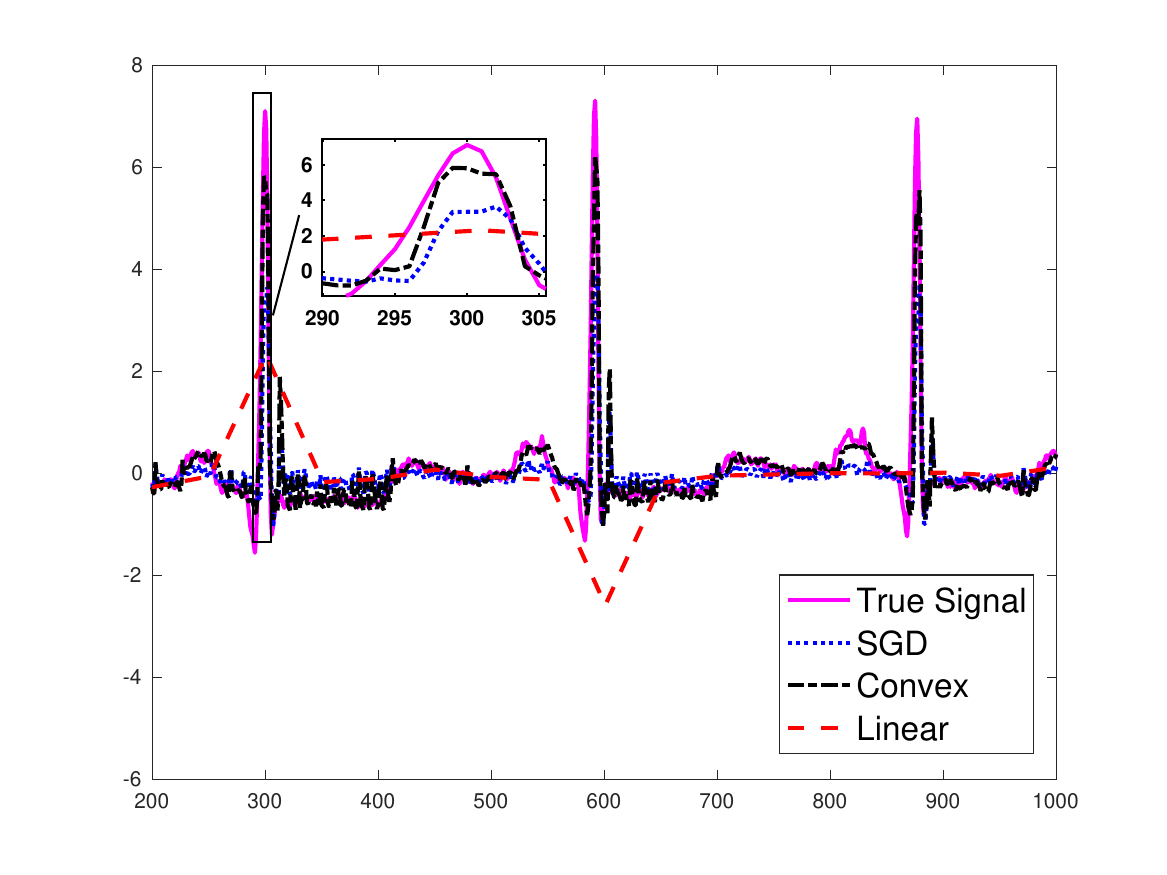}
 x       \caption{Linear versus network predictions}
        \label{fig:ecg_nonlin}
    \end{subfigure}
    \hspace{1.5cm}
    \begin{subfigure}[t]{0.35\textwidth}
        \centering
        \includegraphics[width=1\textwidth]
        {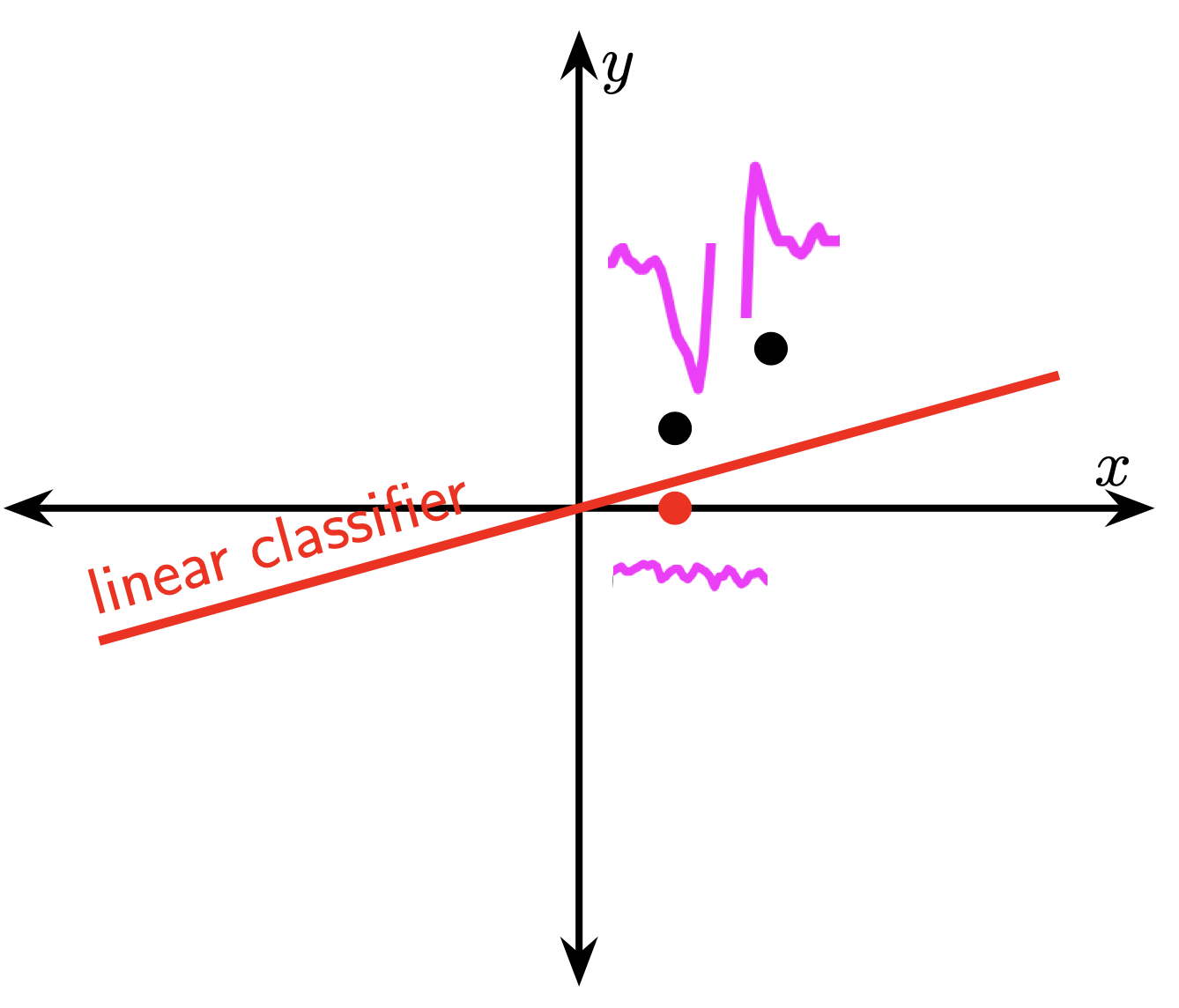} 
         \captionsetup{width=1.1\textwidth}
        \caption{\minorrevision{NN}s are implicit local linear classifiers.}
        \label{fig:ecg_why}
    \end{subfigure}    
     \caption{Two-layer ReLU \finalrevision{NN} trained with SGD \finalrevision{versus} convex model on ECG data (reproduced from \cite{Ergen2023Landsape}).  \finalrevision{Convex training} achieves lower training loss \eqref{fig:ecg_obj} and better test performance \eqref{fig:ecg_pred} compared to SGD. 
     The convex \finalrevision{model} shows that \minorrevision{NNs} are local linear models that can adapt to different data regions 
     \eqref{fig:ecg_nonlin}, \eqref{fig:ecg_why}.}
     \label{fig:ecg_results}
 \end{figure}

\minorrevision{In addition to theoretical guarantees,} simulations demonstrate performance advantages of training \minorrevision{NN}s by using their convex models.
    \revision{In \cite{gupta2021convex}, networks trained on convex programs from \cite{pilanci2020neural} are shown to perform better or comparably to networks trained on \nc{} training problems for standard autoregressive benchmarking tests including MNIST.} We focus on two additional experiments using the convex approach in \cite{pilanci2020neural} \finalrevision{for autoregression}.
%
Autoregression \minorrevision{models} a signal $\mathbf{x} \in \mathbb{R}^d$ at time $t$ as a function of its past $T$ samples $\mathbf{x}_t = f_\theta(\mathbf{x}_{t-1},\cdots,\mathbf{x}_{t-T}) $
where $\theta$ is a parameter set. A linear autoregressive model is \minorrevision{of the form} $f_\theta(\mathbf{x}_{t-1},\cdots,\mathbf{x}_{t-T})=a_1 \mathbf{x}_{t-1}+\cdots+a_{T}\mathbf{x}_{t-T} $ where the \minorrevision{set of} coefficients $\theta =\{ a_1, \cdots, a_T\}$ is estimated from the data. Alternatively, \minorrevision{a NN can replace} the linear model $f_\theta$ \minorrevision{to increase expressivity of the predictor}. 
The following two experiments on financial and ECG data investigate \minorrevision{the use of} convex models to train \minorrevision{two-layer NNs for autoregression}.\\ 
The first experiment performs time series forecasting on the New York stock exchange dataset \cite{james2023ISLP}. Non-linear autoregressive models predict the log volume of exchange from the log volume, Dow Jones return, and log volatility in the past $5$ time steps. \revision{The models use 4276 training samples and 1770 validation samples.} \Cref{fig:nyse} illustrates the results, \minorrevision{plotting the mean squared error (MSE) over validation data for different values of $\beta$.}
The \minorrevision{NNs} trained with convex programs (green dashed line) \minorrevision{perform the best overall.} Here, a custom proximal algorithm was developed to solve the convex program \citep{mishkin2022fast}.
\minorrevision{The experiment compares different optimizers for \nc{} training: \minorrevision{stochastic gradient descent} (SGD) with momentum and Adam.} These \nc{} optimizers result in variability between runs, indicated by the shaded area representing one standard deviation across 5 random initializations. In contrast, the convex solver produces consistent and lower error predictions regardless of the initialization.
The second experiment, from \cite{Ergen2023Landsape}, measures the performance of \minorrevision{NN} training algorithms in a time series prediction task using ECG \minorrevision{voltage} data. \Cref{fig:ecg_results} illustrates results. 
Each \minorrevision{training} sample consists of three consecutive voltage values \minorrevision{$(d=3)$}, \minorrevision{and the model} aims to predict the next voltage value. The dataset contains \( n = 2393 \) observations.
A two-layer ReLU network is trained using both SGD and the convex \lasso{} method. SGD used a batch size of 100 and a grid of learning rates.
 As shown in Fig.~\ref{fig:ecg_results}, the convex optimization method outperforms SGD in both training loss and test prediction accuracy.
 The convex method, which yields a globally optimal NN, obtains a lower training objective than SGD, which can get stuck in local minima. The figure shows that SGD demonstrates poorer generalization,
 demonstrating practical advantages of the convex approach in signal processing applications. 
 The \lasso{} formulation also reveals why \minorrevision{NN}s perform better than linear methods. In contrast to a linear classifier which treats all signals \minorrevision{with the same model}, whether they are spiking or in a rest phase, the \lasso{} \minorrevision{model} acts as a \minorrevision{local} linear classifier, treating different types of signals with different \minorrevision{models}, as visualized in  \Cref{fig:ecg_results} (d). 

\minorrevision{
The convex models \finalrevision{naturally admit the} application of specialized solvers used in signal processing such as proximal gradient methods, which are used in total variation denoising, audio recovery, and other applications \cite{Antonello2018}. \finalrevision{Although \nc{} optimization can also use these methods, convex settings provide advantages such as global guarantees and certificates of optimality via duality and principled stopping
criteria.}
}
 These solvers achieve faster convergence and robustness compared to traditional methods \cite{mishkin2022fast}. \Cref{fig:convexnn} from \cite{mishkin2022fast} compares the proportion of problems solved to $10^{-3}$ relative training accuracy over 400 UC Irvine datasets. The blue and red curves in \Cref{fig:convexnn} plot timing profiles for the convex problem, \minorrevision{comparing a custom proximal solver \cite{mishkin2022fast} and MOSEK. SGD and ADAM are applied to the \nc{} objective, and the experiment selects the best \minorrevision{NN} over a grid of 3 random initializations and 7 step sizes.} Convex \finalrevision{training} achieves a global \minorrevision{optimum} and \minorrevision{better training performance} than \nc{} training.



\section*{\minorrevision{Conclusion}}
The training \minorrevision{problem} of \minorrevision{a} two-layer ReLU \minorrevision{NN} can be equivalently formulated as a convex optimization problem \revision{in a variety of ways, including} within the frameworks of \lasso{} and \minorrevision{g}roup \lasso{} from sparse signal processing and \minorrevision{CS}. \minorrevision{Networks can be expressed as convex programs with respect to the outermost weights \cite{bengio2005convex}, network outputs \cite{Haeffele2017}, or parameter sets \cite{pilanci2020neural}. Convexification approaches range from treating the weights as probability distributions and the network as an expectation \cite{bach:2022, mei2018mean}, to quantizing weights and formulating the training \minorrevision{problem} as a linear program \cite{BIENSTOCK2023}, to leveraging convex duality with hyperplane arrangements \cite{pilanci2020neural}. Various approaches \cite{bengio2005convex}, \cite{pilanci2020neural} center on a convex program with a $l_1$ penalty on the outermost weights, which can be solved or estimated with techniques related to boosting \cite{bengio2005convex} or \minorrevision{CS} \cite{pilanci2020neural}  \cite{Ergen2023Landsape}. The convex networks can be interpreted geometrically, as polyhedra representing classes of linear programs \cite{BIENSTOCK2023}, or via hyperplane arrangements and zonotope vertices \cite{pilanci2020neural}, \cite{Ergen2023Landsape}, or as high-dimensional volumes using geometric algebra \cite{pilancicomplexity}.}
%
\revision{Convex} approaches offer theoretical insights and practical advantages, including stability, interpretability, efficient optimization and improved generalization. \minorrevision{The convex programs for regularized training also overcome NP-hard complexities of unregularized training.} Future work includes extending these methods to deeper networks with modern activations and exploring their implications in various signal processing and machine learning applications.

%
%

%




\section*{Acknowledgement}
This work was supported in part by the National Science Foundation (NSF) CAREER Award under Grant CCF-2236829, in part by the National Institutes of Health under Grant 1R01AG08950901A1, in part by the Office of Naval Research under Grant N00014-24-1-2164, and in part by the Defense Advanced Research Projects Agency under Grant HR00112490441. Emi Zeger was supported in part by in part by the National Science Foundation Graduate Research Fellowship under Grant No. DGE-1656518.

%
%

 %
\end{document}